\pdfoutput=1
\documentclass[11pt]{article}

\usepackage[final]{acl}

\usepackage{times}
\usepackage{latexsym}

\usepackage[T1]{fontenc}
\usepackage[utf8]{inputenc}

\usepackage{microtype}

\usepackage{inconsolata}

\usepackage{graphicx}
\usepackage{amsmath}
\usepackage{xspace}
\usepackage{subfigure}
\usepackage{multirow}
\usepackage{cleveref}
\usepackage{amssymb}
\usepackage{booktabs} 
\usepackage{url}
\usepackage{lscape}
\usepackage{afterpage}
\usepackage{rotating}
\usepackage{ulem}
\usepackage{diagbox}
\usepackage{listings}
\usepackage{hyperref}

\newcommand{\framework}{\texttt{DPT-Agent}\xspace}

\title{Leveraging Dual Process Theory in Language Agent Framework \\for Real-time Simultaneous Human-AI Collaboration}

\author{
 \textbf{Shao Zhang\textsuperscript{1}\thanks{Equal Contribution}},
 \textbf{Xihuai Wang\textsuperscript{1}\footnotemark[1]\thanks{Work done while interning at Meituan.}},
 \textbf{Wenhao Zhang\textsuperscript{1}},
 \textbf{Chaoran Li\textsuperscript{1}}, 
\\
 \textbf{Junru Song\textsuperscript{1}},
 \textbf{Tingyu Li\textsuperscript{1}},
 \textbf{Lin Qiu\textsuperscript{2}},
 \textbf{Xuezhi Cao\textsuperscript{2}},
 \textbf{Xunliang Cai\textsuperscript{2}},
\\
 \textbf{Wen Yao\textsuperscript{3}},
 \textbf{Weinan Zhang\textsuperscript{1}},
 \textbf{Xinbing Wang\textsuperscript{1}},
 \textbf{Ying Wen\textsuperscript{1}\thanks{~Corresponding Author. } }
\\
 \textsuperscript{1}Shanghai Jiao Tong University,
 \textsuperscript{2}Meituan,
 \textsuperscript{3}Intelligent Game and Decision Laboratory
\\
  \texttt{\{shaozhang,leoxhwang,ying.wen\}@sjtu.edu.cn} \\
 }

\begin{document}
\maketitle
\begin{abstract}

Agents built on large language models (LLMs) have excelled in turn-by-turn human-AI collaboration but struggle with simultaneous tasks requiring real-time interaction. Latency issues and the challenge of inferring variable human strategies hinder their ability to make autonomous decisions without explicit instructions.
Through experiments with current independent \textit{System 1} and \textit{System 2} methods, we validate the necessity of using Dual Process Theory (DPT) in real-time tasks.
We propose \textbf{\framework}, a novel language agent framework that integrates \textit{System 1} and \textit{System 2} for efficient real-time simultaneous human-AI collaboration.
\framework's \textit{System 1} uses a Finite-state Machine (FSM) and code-as-policy for fast, intuitive, and controllable decision-making. 
\framework's \textit{System 2} integrates Theory of Mind (ToM) and asynchronous reflection to infer human intentions and perform reasoning-based autonomous decisions.
% We demonstrate the effectiveness of \framework through further experiments\footnote{The up-to-date evaluation results will be maintained in \url{https://agi-eval.cn/evaluation/Evaluation?id=56}.} with rule-based agents and human collaborators, showing significant improvements over mainstream LLM-based frameworks.
We demonstrate the effectiveness of \framework through further experiments\footnote{The up-to-date evaluation results will be maintained in \href{https://agi-eval.cn/evaluation/Evaluation?id=56}{AGI-Eval: Overcooked Challenge}.} with rule-based agents and human collaborators, showing significant improvements over mainstream LLM-based frameworks.
\framework can effectively help LLMs convert correct slow thinking and reasoning into executable actions, thereby improving performance.
To the best of our knowledge, \framework is the first language agent framework that achieves successful real-time simultaneous human-AI collaboration autonomously. 
Code of \framework can be found in \url{https://github.com/sjtu-marl/DPT-Agent}.
\end{abstract}

\section{Introduction}

Large language models (LLMs) have revolutionized generalization capabilities and interaction methods, driving the application of human-AI collaboration in real-world tasks.
LLM-based agents have already been successfully applied to many collaborative tasks with humans, such as writing \cite{wan2024felt} and coding \cite{prather2024widening}, where humans and the agents interact turn-by-turn.
However, many collaborative tasks in shared workspaces require entities involved in the collaboration to cooperate simultaneously in the environment \cite{vildan2024auto,collaborationawareness1992}. 
Unlike turn-by-turn collaborative tasks, the simultaneous collaboration tasks that are time-sensitive require real-time responses to partners and interaction with the environment \cite{shao2024collaborative,gong2024mindagent}, as well as reasoning about dynamically changing human partners' strategies and environments \cite{wang2024zsc}.
Such simultaneous human-AI collaboration tasks present two challenges for LLM-based agents: \textbf{real-time responsiveness and autonomous collaboration adapted to humans}.

\begin{figure}[t]
    \centering
    \includegraphics[width=\linewidth]{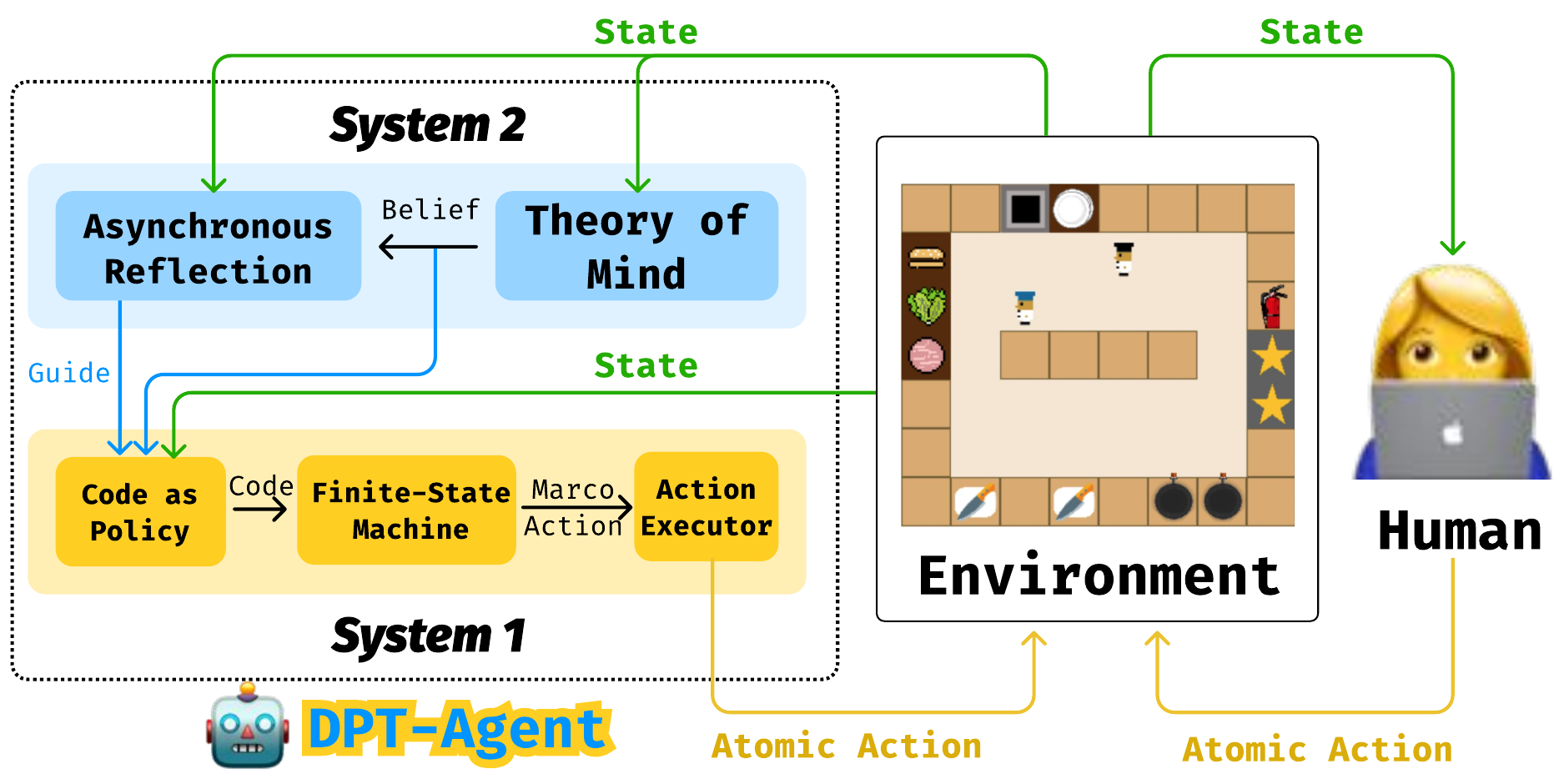}
    \caption{How \framework Collaborates with Human Simultaneously.}
    \label{fig:intro}
\end{figure}

The real-time responsiveness issues faced by LLMs in inference time have been widely discussed.
Larger models with stronger reasoning capabilities often suffer from significant latency \cite{zhou2024survey}, making it difficult for them to respond quickly to dynamic changes in human interactions and environments in highly real-time scenarios.
The combination of fast and slow thinking using \textit{System 1} and \textit{System 2} based on Dual Process Theory (DPT) \cite{kahneman2011thinking,evans2013dual} has already been applied to address real-time issues via the combination of large and small models in language agent frameworks \cite{liu2024slow}.
However, this method still cannot resolve the contradiction between latency and performance fundamentally, as it uses small models as \textit{System 1}.

The agent frameworks designed for collaborating with humans also face challenges of insufficient autonomy and difficulty in adapting to human strategy variability.
Agents in the shared workspace tasks are regarded as independent collaborators joining the partnership \cite{dafoe2021cooperative}. 
However, most collaborative agent frameworks still require human input to output actions or strategies \cite{liu2024slow,guan2023efficient}, failing to collaborate with humans autonomously. 
Furthermore, humans in shared workspace tasks might perceive and engage with agents like how they interact with human partners for fostering collaboration like inferring agents' intentions to adjust strategies \cite{zhang2024mutual}, which further enhances the challenge of simultaneous human-AI collaboration.
Researchers also point out that LLMs are still limited in their ability to adapt to dynamic human strategy changes \cite{DBLP:conf/acl/ZhangTWW0HTLZ024}, making it difficult to transition reasoning into decision-making for effective adaptation \cite{riemer2024can}.

To address these challenges, we propose \textbf{\framework}, which leverages Dual Process Theory (DPT) to integrate FSM-based \textit{System 1} and LLM-driven \textit{System 2}, as shown in \Cref{fig:intro}.
Based on the intuitive thinking and fast decision-making characteristics of \textit{System 1}, we use a Finite-state Machine (FSM) for low-level action decision-making and execution, while employing a code-as-policy \cite{DBLP:conf/icra/LiangHXXHIFZ23} approach to enable \textit{System 2}'s slow thinking to guide and control fast decisions.
For slow thinking (\textit{System 2}), we design a Theory of Mind (ToM) mechanism for actively inferring human intentions and reflecting on environmental feedback based on how humans infer the partners and situations in shared workspace collaboration \cite{krych2007think}.
We also further improve the performance of the reflection mechanism with an asynchronous design to achieve better efficiency in self-evolution.

Building on the shared workspace task environment which is a hard version of Overcooked from \citet{zhang2024mutual}, we further develop a real-time simultaneous human-AI collaboration environment with new layouts and conduct multiple experiments in single agent setup, with rule-based agent and real humans.
We aim to understand:
1) \framework's capability in real-time tasks, 2) \framework's capability in collaboration, and 3) \framework's performance in collaboration with humans simultaneously.

In the experiments collaborating with rule-based agents, \framework outperforms strong language agent frameworks.
In reasoning models that suffer from extremely high latency due to long thinking processes, \framework framework further demonstrates its ability to effectively convert thinking into action and improve performance.
When collaborating with real humans, \framework also outperforms these baselines in both subjective and objective results, showing the significant improvement brought by asynchronous reflection and ToM module to infer humans. 

In summary, our contributions are as follows: 
\begin{itemize}
\vspace{-5pt}
    \item We experimentally analyze LLMs independently as \textit{System 1} and \textit{System 2} in real-time tasks, highlighting the challenge of the trade-off between performance and latency.
    \vspace{-5pt}
    \item We propose \framework that integrates FSM-based \textit{System 1} for fast and intuitive decision-making and LLM-driven \textit{System 2} for deliberate and analytical reasoning, effectively balancing latency and performance.
    \vspace{-5pt}
    \item We conduct extensive experiments with rule-based agents and human participants, demonstrating that \framework outperforms existing language agent frameworks in real-time simultaneous human-AI collaboration.
\end{itemize}

To the best of our knowledge, \framework is the first agent framework that can achieve successful real-time simultaneous human-AI collaboration autonomously in the hard version of Overcooked, which is one step closer to real-world application.

\section{Related Works}

\paragraph{Dual Process Theory (DPT).}
Dual Process Theory (DPT) \cite{evans2013dual} refers to human cognition operates through two distinct systems: \textit{System 1}, which is fast, automatic, and intuitive, and \textit{System 2}, which is slower, deliberate, and analytical \cite{kahneman2011thinking}.
DPT explains how humans think during the perception-decision process. 
The ability to effectively integrate \textit{System 1} and \textit{System 2} helps humans accomplish complex perception and decision-making tasks.
Numerous LLM-based reasoning frameworks also utilized DPT to facilitate human-related interactions like dialogue \cite{he-etal-2024-planning} and mitigate latency issues via using a small model as \textit{System 1} \cite{liu2024slow}.
Many current agent frameworks use \textit{System 2}-based approaches to assist with planning and decision-making \cite{yu2024distilling,DBLP:conf/acl/ZhangTWW0HTLZ024}, such as chain-of-thought (CoT) \cite{wei2022chain}, ReAct \cite{yao2022react}, and Reflexion \cite{shinn2024reflexion}.
\framework is inspired by DPT, further alleviating latency issues in \textit{System 1} and endowing the agent with greater autonomy and adaptability to humans in the design of \textit{System 2}.

\paragraph{Simultaneous Human-AI Collaboration.}
Most tasks related to LLMs in human-AI collaboration research pose lower demands on real-time responsiveness, such as task-oriented dialogue systems \cite{yi2024survey} and word-guessing \cite{zahra2021direction}, where players take actions turn-by-turn.
However, collaborative tasks in the real world are often simultaneous, requiring real-time reasoning, which presents latency challenges for many LLM-based frameworks \cite{DBLP:conf/icra/LiangHXXHIFZ23}.
Another significant challenge of simultaneous collaborative tasks is adapting to humans, who are unfamiliar partners not encountered during training \cite{wang2024zsc,Yang23Cole,carroll2019utility,zhang2024proagent}. 
Theory of Mind (ToM) \cite{neil2018tom,baron1985does} has been introduced to enhance reasoning in human-AI collaborative scenarios \cite{wester2024theory}.
However, studies have pointed out that LLMs fail to achieve functional ToM \cite{riemer2024can}, where reasoning cannot be effectively implemented in decision-making processes.
To adapt to humans, \framework integrates DPT and ToM to support the entire process from perception to reasoning and decision-making, achieving functional ToM while ensuring real-time performance.

\section{Why We Need Dual Process Theory?}

To understand the necessity of DPT in real-time simultaneous human-AI collaboration, we first examine the real-time responsiveness and task completion capabilities of using large language models (LLMs) independently as \textit{System 1} and \textit{System 2} agents.

In the Overcooked environment \citep{zhang2024mutual}, we employ a single-agent setup, \textit{Counter Circuit} (shown on the left in \Cref{fig:overcookedlayout}), to compare the performance of typical LLM-based \textit{System 1}-only agents using mainstream LLMs of varying sizes with that of an FSM-based agent. Additionally, we include the DeepSeek-R1 series reasoning model \cite{guo2025deepseek} and OpenAI's o3-mini, which incorporate \textit{System 2} capabilities with long CoT as agents for this task.

To evaluate the performance of the \textit{System 1}-only agents in real-time task completion, we assess action output latency, task score, and score efficiency. Each model is evaluated over 20 runs, using the same game introduction prompt (\Cref{app:gameprompt}), instruction prompt (\Cref{app:instruct}), and output prompt (\Cref{app:act-prompt}).

\begin{figure}
    \centering
    \includegraphics[width=\linewidth]{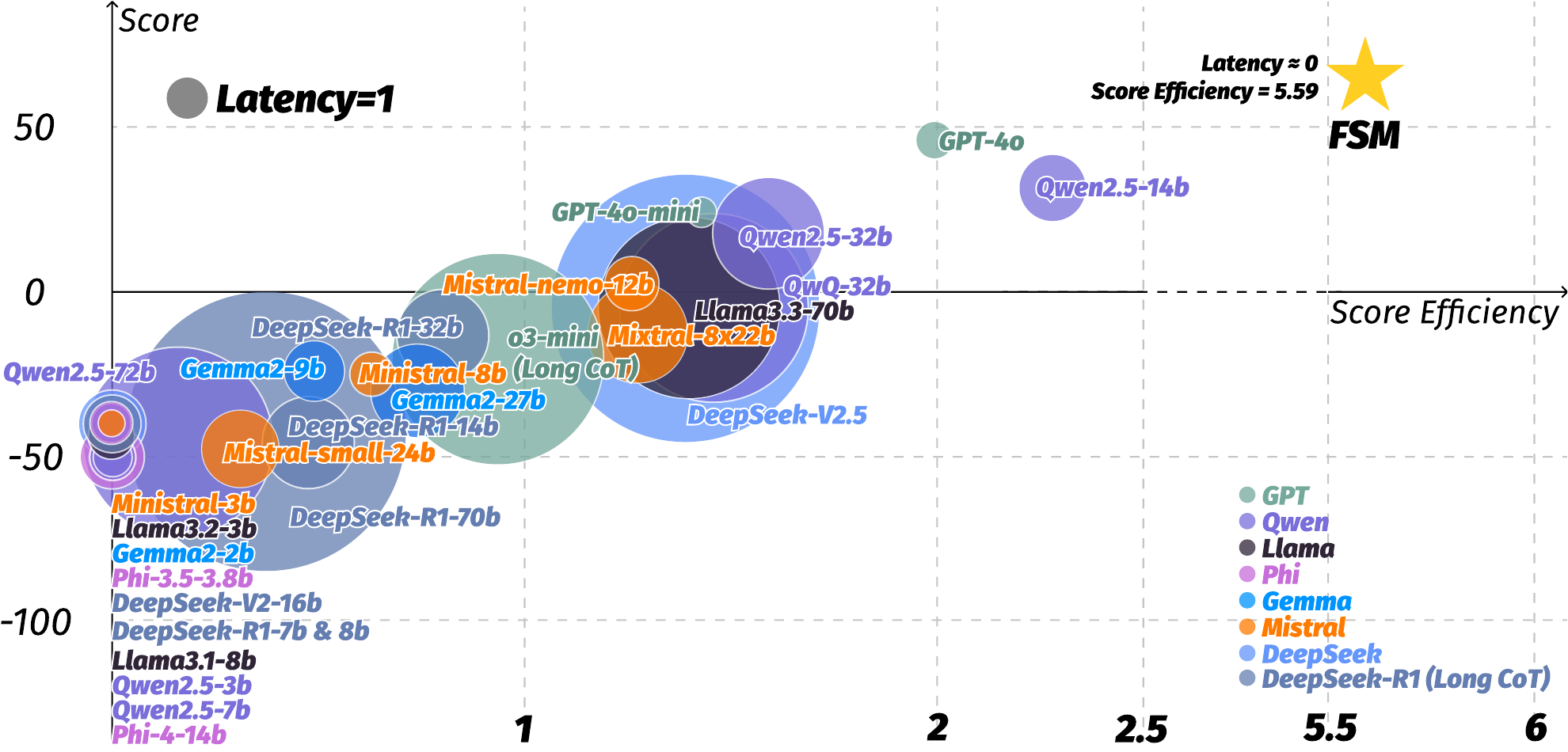}
    \caption{\textbf{LLM as Independent \textit{System 1} and \textit{System 2} in Overcooked.} Mean score means the inter-quartile mean score of 20 games. We define score efficiency as the average score gained per macro action. The size of each model's circle represents latency, which is the time taken from the request to the output of a macro action. }
    \label{fig:exp1}
\end{figure}

As shown in \Cref{fig:exp1}, with detailed data provided in \Cref{app:exp1}, as independent \textit{System 1}, models with fewer than 20B parameters excel in latency but often have near-zero score efficiency, indicating fast responses but ineffective actions.
Since missed orders lead to score deductions, some high-score-efficiency models with high latency still score below zero.
The models that can balance capability in generating scoring actions with low latency perform better.
When the reasoning models use long CoT as the \textit{System 2}, despite their stronger reasoning capabilities, their performance presents even lower score efficiency and overall scores compared to many smaller models functioning as \textit{System 1}.
Additionally, all agents perform worse than the FSM agent.

These results show that LLM-based independent \textit{System 1} and \textit{System 2} agents struggle with low-latency models lacking capability and high-capability models suffering from excessive latency.
This phenomenon highlights the need for a framework to integrate \textit{System 1} and \textit{System 2}, balancing capability and latency in real-time tasks.

\section{\framework Framework}
To enable real-time responsiveness and seamless autonomous collaboration that aligns with human cognitive processes, we propose \textbf{D}ual \textbf{P}rocess \textbf{T}heory \textbf{Agent} framework (\textbf{\framework}). 
\framework integrates both \textit{System 1}, which facilitates fast, intuitive decision-making, and \textit{System 2}, which supports deliberate, analytical reasoning. 

\paragraph{Formulation.} We model real-time simultaneous human-AI collaboration as a two-agent decentralized Markov decision process (DEC-MDP) \cite{bernstein2002complexity}. The framework is defined by the tuple $\langle\mathcal{S}, \{\mathcal{A}^i\}, \{\mathcal{A}^h\}, \rho, \mathcal{P}, r\rangle$ where $\mathcal{S}$ is the state space, $\mathcal{A}^i$ and $\mathcal{A}^h$ denote the agent's and human's action spaces, $\rho:\mathcal{S}\to[0,1]$ is the initial state distribution, $\mathcal{P}:\mathcal{S}\times\mathcal{A}\times\mathcal{S}\to[0,1]$ governs transitions with $\mathcal{A} = \mathcal{A}^i \times \mathcal{A}^h$ as the joint action space, and $r:\mathcal{S}\times\mathcal{A}\to\mathbb{R}$ is the reward function. At each timestep $t$, the agent executes $a^i_t \in \mathcal{A}^i$ while the human performs $a^h_t \in \mathcal{A}^h$ simultaneously, inducing the joint action $a_t = (a^i_t, a^h_t)$ that drives state transitions through $\mathcal{P}(s_{t+1}|s_t, a_t)$. We further develop modular formulations for \framework in the following sections.

\begin{figure*}
\vspace{-10pt}
    \centering
    \includegraphics[width=\linewidth]{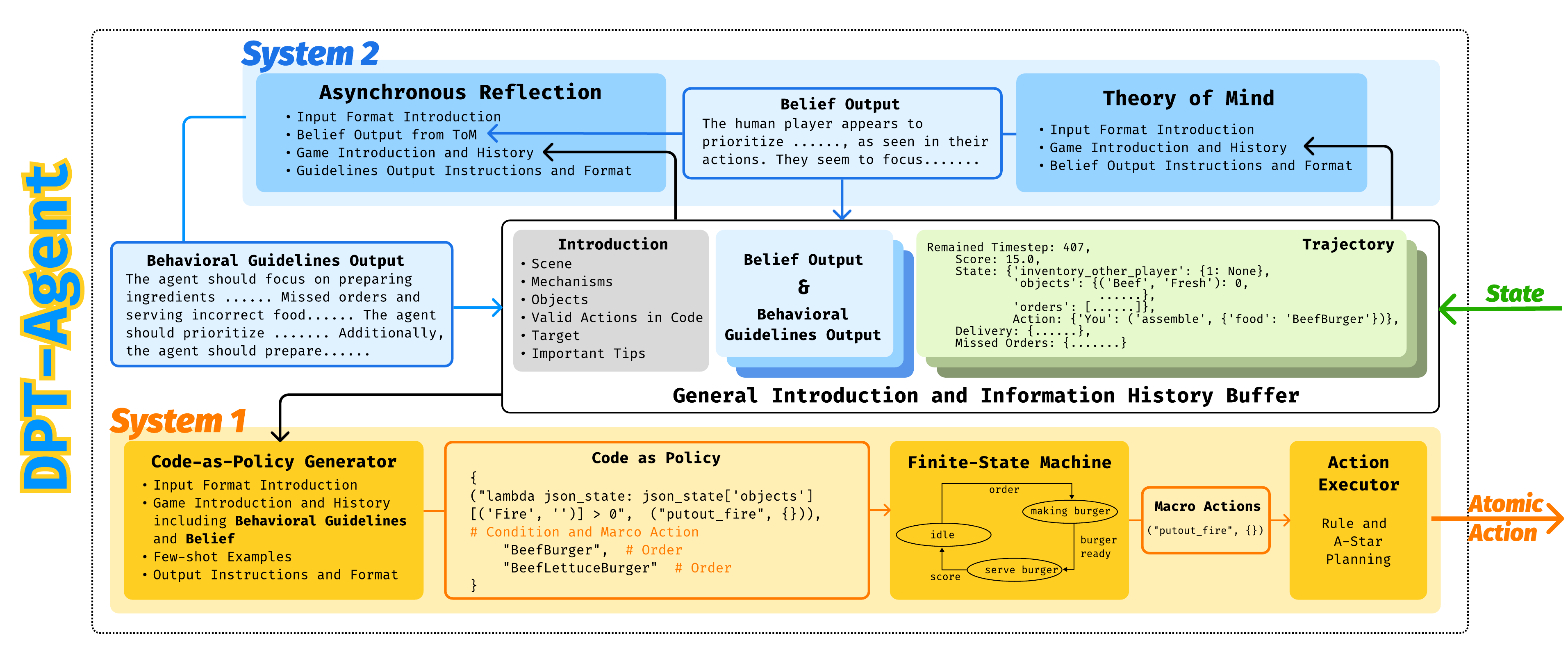}
    \caption{\textbf{\framework Framework.} 
 In \textit{System 2}, the historical states from the history buffer periodically trigger the ToM module to infer human behaviors. The reflection module then analyzes the belief output from the ToM module, along with game score feedback and other historical state information, to summarize its own behaviors and generate guidelines. 
Within \textit{System 1}, the code-as-policy generator utilizes the current state, belief and guidelines to generate code-as-policy when necessary, enabling control over the FSM. When no specific input is provided, the FSM continues operating autonomously, generating macro actions to ensure the agent maintains continuous action output, thereby guaranteeing real-time responsiveness in simultaneous collaboration.}
    \label{fig:framework}
\end{figure*}

\subsection{\textit{System 2}: Deliberate and Analytical Reasoning}

When facing complex situations, humans often rely on \textit{System 2} to process large amounts of information to aid decision-making. 
Inspired by this process, we designed \textit{System 2} for \framework, integrating environmental feedback for Theory of Mind and self-evolution-based inference, which aims to enable advanced reasoning and planning while dynamically adapting to human partners.
We also refine the reflection mechanism \cite{shinn2024reflexion} by using asynchronous reflection to facilitate efficient and flexible self-evolution of strategies.

\subsubsection{Theory of Mind for Inferring Human }

Equipped with the Theory of Mind (ToM) capability, individuals can infer others' mental states as beliefs by analyzing their actions and communication history, allowing them to understand and anticipate their behaviors \cite{premack1978does}.
In the context of ToM, belief refers to an individual's perception of events, which subsequently shapes their actions \cite{baron1985does,neil2018tom,wen2018probabilistic}.
We develop a Theory of Mind module that enables the agent to construct a belief about the human, encompassing aspects such as tendencies, conventions, and plans, based on observed human behaviors.
The belief output from the ToM module influences the strategy by guiding both the strategy reflection in \textit{System 2} and the decision-making in \textit{System 1}.

To formulate the ToM process, we denote the history from time-step $0$ to time-step $t$ of the game that the agent perceives as $\tau_{0:t} = \{(s_{0}, a_{0}^{i}, a_{0}^{h}, r_{0}), \ldots, (s_{t}, a_{t}^{i}, a_{t}^{h}, r_{t})
\}$.
The Theory of Mind module takes in the history $\tau_{0:t}$, summarizes the history, infers the conventions and tendencies of the human, and explains how the agent's policy can be adjusted to coordinate better with the human player. 
The Theory of Mind module outputs the belief in natural language, as shown in \Cref{fig:framework}. 
The $n$-th ToM process execution can be formalized as $b^{n} = \text{LLM}\left(\tau_{0:t_{n}}, b^{n-1}\right)$, where $b^{n-1}$ is the $n-1$-th generated belief and $t_{n}$ is the time-step when the $n$-th belief inference is performed.

\subsubsection{Asynchronous Reflection for Self-evolution}
The Asynchronous Reflection module enables the agent to improve its policy in such a long-horizon interaction process for higher performance.
We design the ``Behavior Guideline,'' where the agent maintains and iteratively updates language guidelines for the self-evolution of the current policy, based on the generated belief about the human partner and the game history.
The Asynchronous Reflection module proceeds asynchronously with the decision-making process and allows real-time responsiveness to be handled by \textit{System 1}, enabling the reflection process to focus on thinking without worrying about decision delays, thus facilitating more thorough self-evolution.
The $m$-th Reflection process execution can be formalized as $g^{m} = \text{LLM}\left(\tau_{0:t_{m}}, b^{n},g^{m-1}\right)$, where $b^{n}$ is the latest inferred belief about human, $g^{m}$ is the ``Behavior Guideline'' that is updated $m$ times.

Given the modular formulation of the ToM and Asynchronous Reflection modules, we derive the formulation of the whole \textit{System 2} process as a policy $\pi^{\text{S2}}: \mathcal{T} \times \mathcal{B} \times \mathcal{G} \mapsto \mathcal{B} \times \mathcal{G}$, where $\mathcal{T} = \left\{ \tau_{0:t} = (s_0, a_0, \dots) \mid s_t \in \mathcal{S}, a_t \in \mathcal{A}, t = 0, \dots \right\}$ is the space of the game history. 
The \textit{System 2} policy $\pi^{\text{S2}}$ iteratively updates the belief about the human player and the behavior guidelines given the game history, which can be denoted as $b^{n}, g^{m} = \text{LLM}(\tau_{0, \max(t_n, t_m)}, b^{n-1}, g^{m-1})$.

\subsection{\textit{System 1}: Fast and Intuitive Decision Making}

In time-sensitive tasks, humans typically rely on \textit{System 1} to make intuitive decisions without engaging in complex reasoning and keep asynchronous reasoning while taking action. 
Inspired by this process, we implement \textit{System 1} in \framework by combining a code-as-policy generator and Finite-state Machine (FSM)  to enable intuitive and rapid decision-making.
The code-as-policy approach also establishes a decision pipeline from \textit{System 2} to \textit{System 1}, which allows \textit{System 2} to influence and refine actions.

\subsubsection{Code-as-Policy Generator}

To enhance the performance of the agent, we designed the code-as-policy generator to effectively bridge the gap between \textit{System 2}'s guidelines and inferred beliefs, and \textit{System 1}'s rapid decision-making. 
By incorporating \textit{System 2}'s reasoning into the decision pipeline, we ensure that the agent can leverage \textit{System 2}'s reasoning abilities to gradually transform \textit{System 2}'s inferences into actionable decisions within an episode. 

The Code-as-policy generator takes in the history, guidelines and inferred beliefs, and outputs executable code that consists of task-completing rules and modifies the logic of the Finite-state Machine, which is detailed in \cref{sssec: fsm}.
This process allows \textit{System 1} to refine its intuitive responses with thoughts derived from \textit{System 2}, thus enhancing the agent's overall decision-making capabilities in dynamic environments.

The policy generation process of Code-as-policy generator at time-step $t$ can be formalized as $c_{t} = \text{LLM}\left(\tau_{t-\lambda:t}, b^{n}, g^{m}\right)$, where $b^{n}$ and $g^{m}$ represents the latest belief about human and the latest guidelines respectively, and $\lambda$ is the interval the Code-as-policy generator executes.
% which is $25$ in our experiment.

\subsubsection{Finite-state Machine \& Action Executor}\label{sssec: fsm}

To implement rapid response in \textit{system 1}, we adopt the Finite-state Machine (FSM) method \citep{russell2016artificial}, which is a widely used computational model that enables structured and efficient decision-making by transitioning between pre-defined states based on inputs. 
In \framework, we leverage FSM to facilitate fast and intuitive decision-making by defining each state as a specific agent context or situation. 
State transitions are triggered by environment dynamics, allowing the agent to adapt efficiently without relying on external LLM responses. 

When LLM generates code-as-policy, the executable code changes the pre-defined logic of FSM and thus facilitates the adaption to human and performance improvement.
The FSM takes in the code-as-policy and game states, and outputs macro actions, denoted as $ma$, which are high-level combinations of atomic actions for specific targets.
For example, in Overcooked, macro actions include food ingredients preparation, food assembling and food serving.  
The generated macro actions are sent to an action executor for conversion into atomic actions that can be directly executed in the environment. 
The action executor employs script policies, ensuring smooth and efficient execution.
Upon receiving a macro action, the action executor selects an appropriate execution plan and performs path planning to determine the necessary atomic actions using the A* algorithm \cite{hart1968formal}. 
The Detailed design and implementation of the FSM is provided in \Cref{app:fsm}.
These processes can be formalized as $ma_t = \text{FSM}(c_t, s_t)$ and $a_{t}^{i} = \text{Executor}\left(ma_{t}\right)~$.

Given the formulation of these modules, we derive the formulation of the whole \textit{System 1} process as $\pi^{\text{S1}}: \mathcal{S} \times \mathcal{B} \times \mathcal{G}\mapsto \mathcal{A}$.
At time-step $t$, $\pi^{\text{S1}}$ generates executable atomic action $a_t = \text{Executor}(\text{FSM}(\text{LLM}(\tau_{t-\lambda}, b^{n}, g^{m}), s_t))$.

\section{Experimental Design}

In this section, we introduce the new real-time simultaneous human-AI collaboration environment and tasks we designed based on \citet{zhang2024mutual} and our experimental setup.
Specifically, we aim to understand: 1) \framework's capability in real-time tasks, 2) \framework's capability in collaboration, and 3) \framework's performance when collaborating with humans simultaneously.

\subsection{Overcooked Challenge for Real-time Simultaneous Human-AI Collaboration}
\begin{figure}
    \centering
    \includegraphics[width=\linewidth]{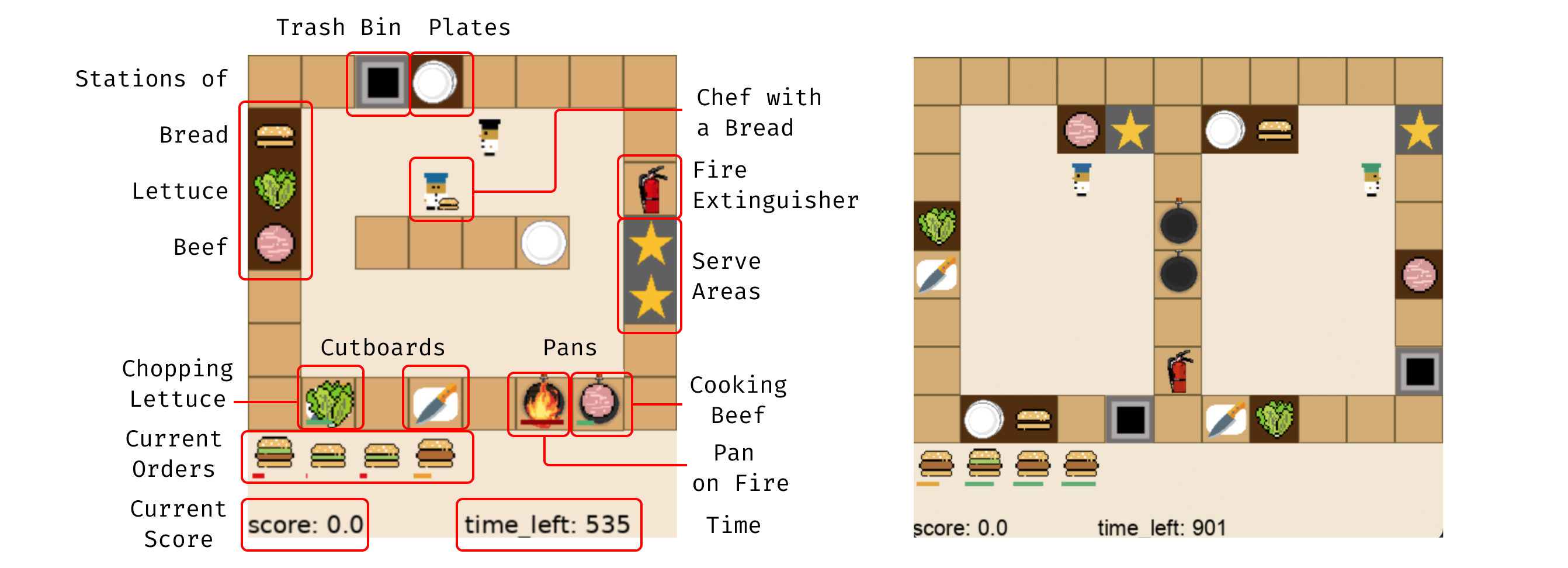}
    \caption{\textbf{Two Layouts in Overcooked Challenge for Real-time Simultaneous Human-AI Collaboration.} Left is Map 1 - \textit{New Counter Circuit} with brief introduction of the item and game mechanism. Right is Map 2 - \textit{New Asymmetric Advantages} }
    \label{fig:overcookedlayout}
\end{figure}

To effectively evaluate the performance of \framework in real-time simultaneous human-AI collaboration, we implement the real-time shared workspace environment proposed by \citet{zhang2024mutual}, using a challenging version of Overcooked based on the original Overcooked game \cite{carroll2019utility,strouse2021fcp,Yang23Cole,li2024tackling,yu23hsp,gymcooking}. In our experiments, we introduce a new layout.
As shown in \Cref{fig:overcookedlayout}, we adopt the basic layout, referred to as \textit{New Counter Circuit}, from \citet{zhang2024mutual} and design a new layout, named \textit{New Asymmetric Advantages}, building on the original Overcooked AI environment \cite{carroll2019utility}. The implementation is based on the gym-cooking environment \cite{gymcooking}.
In the real-time settings, each timestep corresponds to 0.25 seconds in the real world. 
Time-sensitive elements within the environment, such as overcooked beef and expiring orders, underscore the importance of timely task execution. 
Additionally, layout conflicts and the complexity of the burger-making process emphasize the critical role of collaboration. Further details about the environment and tasks can be found in \Cref{app:env}.

\begin{figure*}[htp]
    \centering
    \subfigure[ReAct.]{
    \includegraphics[height=0.22\linewidth]{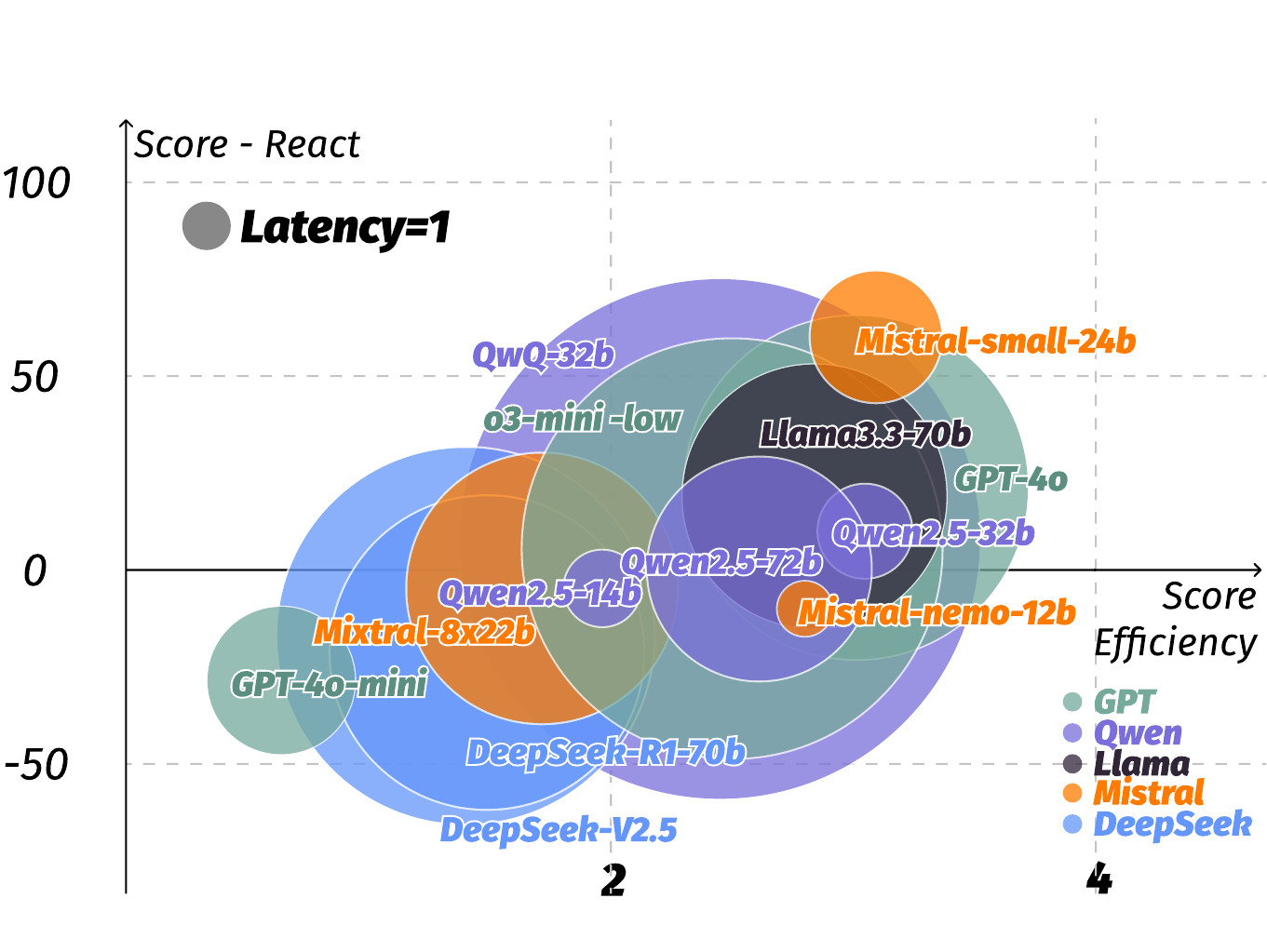}
    \label{fig:exp1-2react}
    }
    \subfigure[Reflexion.]{
    \includegraphics[height=0.22\linewidth]{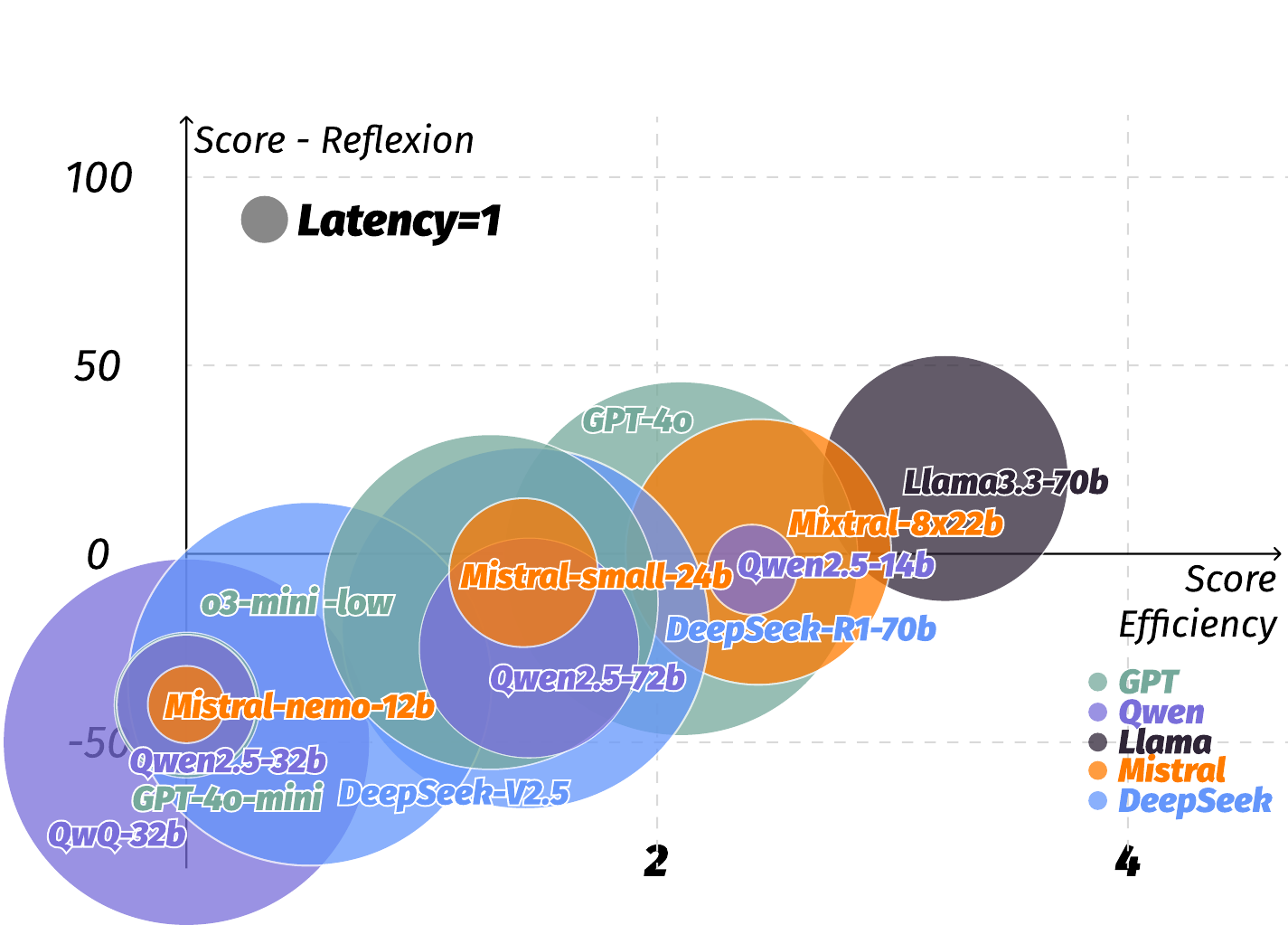}
    \label{fig:exp1-2refexion}
    }
    \subfigure[\framework w/o ToM.]{
    \includegraphics[height=0.22\linewidth]{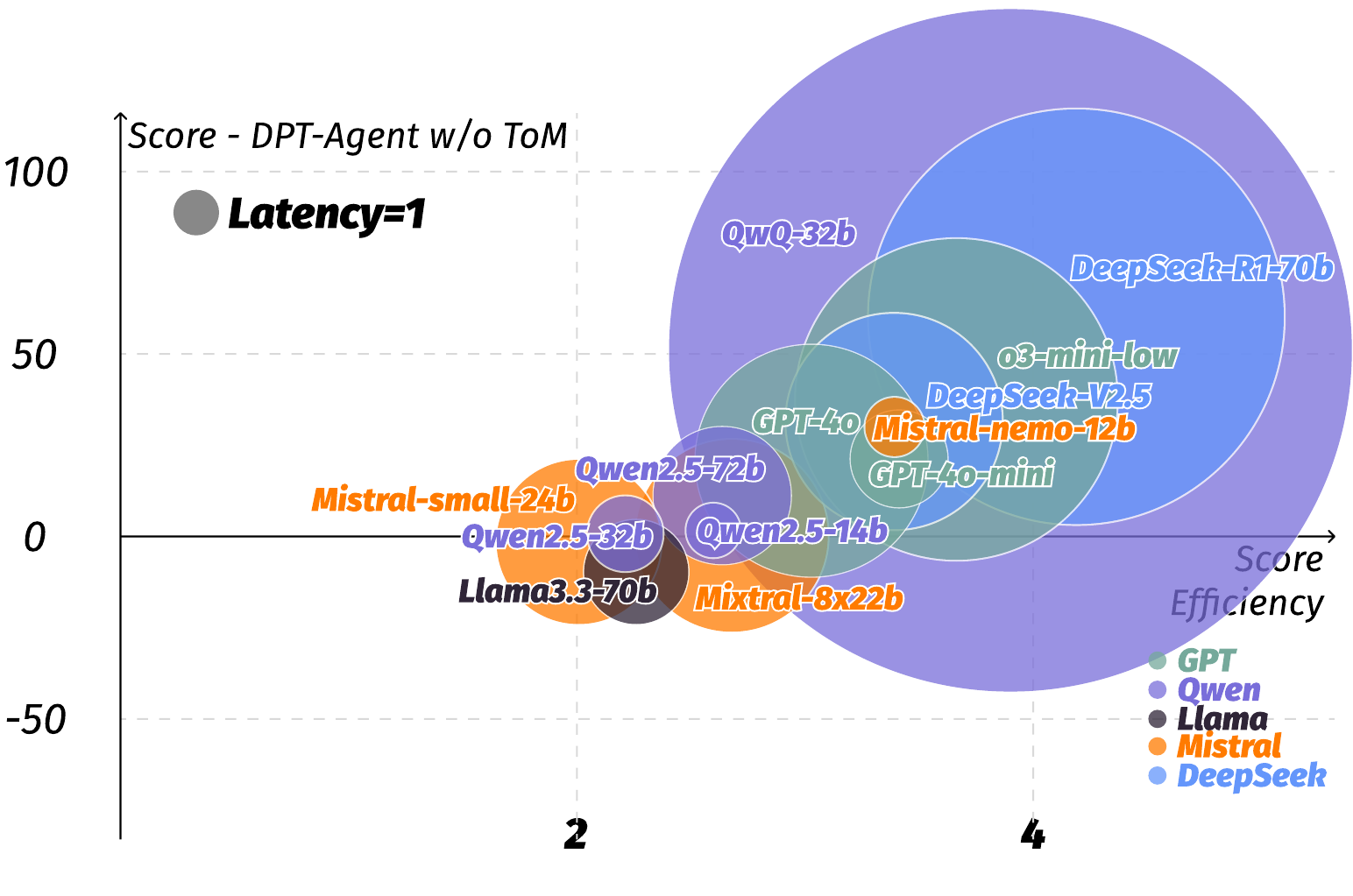}
    \label{fig:exp1-2dpt}
    }
    \caption{\textbf{Results of LLM with ReAct, Reflexion and DPT-Agent w/o ToM in the Single Agent Game.} }
    \label{fig:exp1-2}
\end{figure*}

\subsection{Experimental Setup}

Based on the Overcooked challenge, we set up three series of experiments to validate the effectiveness of \framework using the commonly adopted ReAct \cite{yao2022react} and Reflexion \cite{shinn2024reflexion} framework.
We first compare \framework with baselines in a single agent setting to understand the \framework's capability of a real-time task.
Next, we use three specialized rule-based agents as partners to evaluate the simultaneous collaboration capability of \framework. 
Finally, we conduct human-involved experiments to compare baseline frameworks with \framework in collaboration with real humans.
The baseline frameworks in experiments are implemented in a manner that ensures a fair comparison via using the same output way of code-as-policy with \framework.
Based on this implementation, the ReAct and Reflexion become \textit{System 1} + \textit{System 2} frameworks.
The implementation details can be found in \Cref{app:imple,app:baselines,app:gameprompt}.
All the open-source models used in experiments are deployed locally with NVIDIA A800-SXM4-80GB and NVIDIA H100-80GB-HBM3 for the best latency performance.
Model deployment details can be found in \Cref{app:exp1,app:exp2}.
For close-source models, we use the original API.
All the models' temperature is set to 0.
The whole experiment cost 517.5 A800 GPU hours, 228 H100 GPU hours and \$735 in API in total.
For reliability, all the experiments are repeated 20 runs and reported as the inter-quartile mean and the standard error.
The details of the metrics used in experiments can be found in \Cref{app:metrics}.

\paragraph{Capability in Real-time Task.}
We first consider the real-time performance and task completion capability of \framework in a single agent setting.
In the single-agent setup \textit{Counter Circuit}, we compare the ReAct and Reflexion implemented as \textit{System 1} + \textit{System 2} frameworks with \framework w/o ToM in score, latency and score efficiency.

\paragraph{Capability in Simultaneous Collaboration.}
Expanding on the previous experiment, we use rule-based agents as partners to evaluate the performance of \framework in simultaneous collaboration tasks. We employ three specialized rule-based agents: one for beef preparation, one for lettuce preparation, and one for burger assembly.
In Map 1, we compare ReAct and Reflexion implemented as \textit{System 1} + \textit{System 2} frameworks and \framework w/o ToM with \framework driven by 11 high-performing LLMs on the same map as the previous experiment in a two-player setting.

\paragraph{Real-time Simultaneous Collaboration Experiments with Human.}
To evaluate \framework's capabilities of collaborating with humans, we conduct experiments with 71 university students. To balance response latency and capability, all frameworks are powered by GPT-4o-mini. We enhance all baselines by incorporating an FSM-based \textit{System 1} and perform an ablation study to assess the ToM module's impact.
We compare ReAct + FSM-based System 1, Reflexion + FSM-based System 1, \framework, and \framework w/o ToM on two cooperative two-player maps. Participants are split into two groups, each playing on a different map. Within each group, every participant plays two games, each lasting 500 timesteps, with each of the four agents in random order.
We also collect subjective human preferences. 
Detailed participant demographics, experiment implementation, instructions, recruitment, and payment information are provided in \Cref{app:humanexp}.

\section{Results}
In this section, we present the results of experiments and analyze \framework's effectiveness in real-time simultaneous human-AI collaboration. 

\subsection{Capability in Real-time Task}

As shown in \Cref{fig:exp1-2react,fig:exp1-2refexion} (detailed data in \Cref{app:exp1}), under ReAct and Reflexion framework, the score efficiency of most models has significantly improved compared with when they functioned as independent \textit{System 1} (\Cref{fig:exp1}).
However, the score of many models has declined with an increase in latency due to more complex \textit{System 2} reasoning. 
Low-latency models, like Qwen2.5-14b, still struggle with capability issues, failing to achieve higher final scores despite good score efficiency. 
Further comparison of the performance of \framework in \Cref{fig:exp1-2dpt} reveals that inference models with high latency and larger models get a significant improvement. \framework can help these high latency models convert the high score efficiency and reasoning capability to scores, which demonstrates the effectiveness of \framework in real-time tasks.

\begin{table*}
\vspace{-5pt}
\resizebox{\linewidth}{!}{
\begin{tabular}{lcccccccc}
\toprule
\multicolumn{1}{c}{\multirow{3}{*}{\diagbox[width=3.5cm]{\textbf{Model}}{\textbf{Framework}}}} &
  \multicolumn{4}{c}{\textbf{Score}} &
  \multicolumn{4}{c}{\textbf{Agent Contribution Rate}} \\
  \cmidrule(lr){2-5}
  \cmidrule(lr){6-9}  
 & \multirow{2}{*}{ReAct} & \multirow{2}{*}{Reflexion} & \framework & \multirow{2}{*}{\framework}     & \multirow{2}{*}{ReAct}   & \multirow{2}{*}{Reflexion}     & \framework    & \multirow{2}{*}{\framework}   \\
                &     &        & w/o ToM   &     &    &      &  w/o ToM   &    \\
\midrule
\textbf{o3-mini-high}    & -43.00(0.93)	&-42.00(0.85)&\textbf{65.83(5.66)}&	\uline{55.17(4.84)} &0.00(0.00)&	0.00(0.00)&	0.68(0.02)&	\textbf{0.72(0.01)}   \\			
\textbf{o3-mini-medium}    &-10.00(6.94)	&4.83(7.63)	&\uline{56.50(7.07)}&	\textbf{60.00(6.07)} &0.60(0.05)&	0.62(0.02)&	0.56(0.04)&	\textbf{0.68(0.03)}   \\
\textbf{o3-mini-low}    & 7.00(7.491)	&33.50(7.06)	&\uline{44.83(9.74)}&	\textbf{51.33(8.67)} &0.60(0.05)&	0.62(0.02)&	0.56(0.04)&	\textbf{0.68(0.03)}   \\
\textbf{GPT-4o }     & 35.67(9.62)&\uline{39.17(8.43)}&18.67(8.50)&\textbf{39.50(8.63) } &0.60(0.02)&	0.61(0.02)&	0.60(0.05)&	\textbf{0.69(0.04) }\\
\textbf{GPT-4o-mini}   & -6.58(5.37)&5.58(7.53)&\uline{50.00(5.27)}&\textbf{52.92(6.34)}  &0.27(0.07)&	0.46(0.06)&	0.66(0.02)&	\textbf{0.67(0.02) } \\
\textbf{Qwen-Max}   &30.50(6.58)	&21.17(6.23)&	\uline{51.50(9.27)}	&\textbf{53.83(7.33)}  &0.59(0.03)&	0.60(0.03)&	0.68(0.04)	&\textbf{0.70(0.03)} \\
\textbf{Claude 3.5 Haiku}     & 29.50(5.63)	&24.83(6.58)	&\textbf{43.17(8.01)}	&\uline{41.50(7.69)}  &0.62(0.04)&	0.58(0.03)&	0.67(0.03)&	\textbf{0.70(0.03)}\\
\midrule
\textbf{DeepSeek-R1-671b}     &20.67(5.47)	&21.00(6.83)&\uline{56.67(5.13)}	&\textbf{74.33(5.33) }& 0.61(0.01)&	0.59(0.01)&\textbf{0.69(0.02)}&	\textbf{0.69(0.01)}\\
\textbf{DeepSeek-R1-70b}     &33.83(6.77)&-2.67(5.98)&\uline{51.00(6.08)}&\textbf{61.50(6.40)} &0.57(0.01)&	0.55(0.05)	&\textbf{0.69(0.02)	}&0.66(0.02)\\
\textbf{DeepSeek-R1-32b}     &37.33(8.51)&23.33(7.46)&\textbf{45.50(6.39)}&\uline{38.83(8.51)} &	0.56(0.02)&	0.53(0.03)	&0.67(0.02)&	\textbf{0.69(0.05)}\\
\textbf{DeepSeek-R1-14b}     &-8.50(3.88)&12.00(8.51)&\uline{40.33(7.73)}&\textbf{43.17(8.54)} &0.52(0.02)	&0.48(0.02)	&0.68(0.03)&\textbf{0.71(0.03)}\\
\textbf{DeepSeek-V3 }  &29.17(8.24)&	33.33(7.76)&	\textbf{70.33(5.28)}	&\uline{61.83(5.86)}&   0.60(0.03)	&0.58(0.02)&	\textbf{0.74(0.01)	}&\textbf{0.74(0.02)}  \\
\textbf{DeepSeek-V2.5}   &	-6.00(5.23)	&12.33(4.83)&\textbf{ 31.50(6.58)}&	\uline{23.50(8.44)}&0.25(0.02)	&0.47(0.04)&\textbf{0.64(0.04)}&	0.60(0.04)	\\
\textbf{QwQ-32b}      & \uline{49.17(7.32)}		&-43.33(4.56)&\textbf{53.17(6.00)}	&47.50(6.59)	 &	0.58(0.03)	&0.00(0.03)	&0.64(0.02)	&\textbf{0.70(0.02)} \\
\textbf{Qwen2.5-72b}      & 15.50(4.69)   &\textbf{52.83(5.68)}&18.67(5.51)&\uline{33.67(5.17)	} &	\textbf{0.75(0.01)}&	0.58(0.01)	&0.67(0.04)&	0.67(0.03)  \\
\textbf{Llama3.3-70b}      & 27.97(5.68)&-15.58(5.28)&\textbf{30.75(3.86)}&\uline{28.08(6.68)}    &0.74(0.03)&	0.54(0.05)&	\textbf{0.85(0.02)}	&0.75(0.05)  \\
\textbf{Mixtral-8x22b}    & 20.17(6.30)	&\uline{24.67(6.07)}&	24.00(6.10)	& \textbf{26.83(5.79)}&  0.54(0.03)&	0.54(0.03)	&\textbf{0.70(0.06)	}&0.60(0.03)  \\
\midrule
\midrule
\textbf{Overall}     & 15.48(6.16)	&11.77(6.31)	&\uline{44.23(6.60)}	&\textbf{46.63(6.88)	}  	&0.52(0.03)&	0.49(0.03)
&0.68(0.03)&	\textbf{0.69(0.03)}	 \\
\bottomrule
\end{tabular}
}
\caption{\textbf{Performance with Standard Errors of Experiments Collaborating with Rule-based Agents.}}\label{tab:exp2}
\vspace{-10pt}
\end{table*}

\begin{table}[htp]
\centering
\resizebox{\linewidth}{!}{
\begin{tabular}{@{}lcccc@{}}
\toprule
\multicolumn{5}{c}{\textbf{Map 1 - \textit{New Counter Circuit}}}                                                                               \\ \midrule
\textbf{Frameworks} & ReAct          & Reflexion         & \framework w/o ToM     & \framework \\ \midrule
\textbf{Mean Score} & 99.03(9.86)  &    97.78(7.23)    &      103.19(7.06)     &   \textbf{111.53(5.42)}       \\
\textbf{Agent CR} & 0.51(0.03)  &    0.53(0.03)    &     0.62(0.02)     &   \textbf{0.62(0.02)}   \\
\bottomrule \toprule
\multicolumn{5}{c}{\textbf{Map 2 - \textit{New Asymmetric Advantages}}}                                                                               \\\midrule
\textbf{Frameworks} &  ReAct          & Reflexion         & \framework w/o ToM     & \framework \\\midrule
\textbf{Mean Score} &115.00(9.28)  &   119.67(10.54)  &      152.03(8.13)     &   \textbf{160.63(7.97)}       \\ 
\textbf{Agent CR} & 0.49(0.04)  &    0.51(0.03)    &     \textbf{0.62(0.02)}     &   0.59(0.03)  \\
\bottomrule
\end{tabular}
}
\caption{\textbf{Performance with Standard Errors of Experiments with Humans.} Agent CR refers to Agent Contribution Rate.} 
\label{tab:exp3}
\vspace{-10pt}
\end{table}

\subsection{Capability in Simultaneous Collaboration}

As shown in \Cref{tab:exp2}, \framework achieved the best performance across the majority of models, especially on the widely recognized general-purpose SOTA models like GPT-4o. 
This phenomenon aligns with the conclusions from the experiments in single-agent settings, where larger models can overcome the latency limitations and achieve better performance with the help of \framework. 
Such performance improvements are more noticeable in the reasoning model series of GPT o3-mini and DeepSeek-R1.
\framework framework can help reasoning models, which require long periods of thinking, overcome the latency and effectively transition from thinking to action.
Additionally, when facing rule-based agents that can only perform a single task, \framework can maintain a high contribution rate.
For some models like Llama3.3-70b, \framework w/o ToM outperforms the complete \framework, which may be closely related to the model's ToM capabilities. 
We provide detailed results and case analyses of different partners in \Cref{app:case-study}.

\subsection{Experiments with Real Humans}

After data validation, we have 68 valid data points in total: 36 of Map 1 and 32 of Map 2.
The data validation details are in \Cref{app:humanexp}. 
As shown in \Cref{tab:exp3}, \framework achieves the highest scores in both Map 1 and Map 2 when collaborating with humans. 
\framework w/o ToM also outperforms ReAct and Reflexion, confirming the effectiveness of asynchronous reflection.
Moreover, the ToM module also brought a significant score improvement in collaborating with humans, confirming that incorporating human belief reasoning into \textit{System 2} can foster better collaboration.
Regarding human perception (\Cref{tab:human-rank}), \framework ranks highest in Map 1, with the most participants recognizing its collaborative abilities. 
Interestingly, in Map 2, \framework w/o ToM surpasses \framework in both cooperation and preference ranking with a higher agent contribution rate, which may refer to the human preference for partners who work more.

\begin{table}[htp]
    % \vspace{-10pt}
    \centering
    \resizebox{\linewidth}{!}{
\begin{tabular}{cccccc}
\toprule
  \multirow{2}{*}{\textbf{Layouts}} & \multirow{2}{*}{\textbf{Perception}} & \multirow{2}{*}{ReAct} & \multirow{2}{*}{Reflexion} & \framework& \multirow{2}{*}{\framework}\\
& &&&w/o ToM &\\
\midrule
\multirow{2}{*}{\textbf{Map 1} }       & Cooperation     & \uline{88} & 79 &86 & \textbf{107} \\
\cmidrule{2-6}
     &Preference &  88   & 80  &  \uline{91}  & \textbf{101}  \\
\midrule
\multirow{2}{*}{\textbf{Map 2} }    & Cooperation            & 65 & 78 & \textbf{94} & \uline{83} \\
\cmidrule{3-6}
 &Preference     &      63  &   73  & \textbf{95}       &  \uline{89}   \\
\bottomrule
\end{tabular}
        }
          
        \caption{\textbf{Borda Count Result of Humans' Perceived Subjective Ranking.} In our Borda count, the first place receives 4 points, and each subsequent rank decreases by one point.}
        \label{tab:human-rank}
  \vspace{-10pt}
\end{table}
\section{Discussion and Future Works}

The experiment results illustrate the complex interplay between latency, capability, and collaboration in real-time tasks. 
\framework shows the capability to address this issue by effectively balancing latency and capability, enabling high-latency models to convert score efficiency and reasoning capability into better outcomes.
Moreover, the significant improvement observed when incorporating ToM into \framework during human collaboration confirms the value of human-like reasoning in enhancing task performance. 
This insight emphasizes the importance of integrating cognitive abilities, like ToM, to optimize human-agent interactions in real-world applications.
Interestingly, the absence of ToM in \framework outperformed the complete \framework in some cases, suggesting the models' lack of ToM capabilities, which might influence the effectiveness of \framework. 
For future work, the integration approach of \framework with FSM holds great potential for integrating LLMs into existing FSMs in the real world, offering the possibility of supporting more simultaneous human-AI collaboration scenarios to achieve stronger capabilities and promote better cooperation.

\section{Conclusion}
In this paper, we propose \framework, a language agent framework for the challenges of real-time responsiveness and autonomous adaption to humans in real-time simultaneous human-AI collaboration tasks. 
Inspired by DPT, \framework combines FSM-based \textit{System 1} for rapid decision-making with a \textit{System 2} driven by LLMs for deeper reasoning.
The single-agent experiments and experiments with rule-based agents highlight that \framework has the capability in real-time tasks and simultaneous collaboration.
Moreover, the performance of \framework in human experiments marks a significant advancement, offering a more autonomous and adaptive framework in simultaneous human-AI collaboration. 
We open-source both the method and the environment to foster future research and advancements in simultaneous human-AI collaboration.
To the best of our knowledge, \framework is the first agent framework to achieve autonomous and simultaneous collaboration with humans in real time, making it a major step forward in language agents for human-AI collaboration.

\section*{Limitations}

\framework has already made breakthrough progress in the task of simultaneous Human-AI collaboration, providing a solid foundation for designing more complex agent frameworks in the future. 
However, \framework still has significant room for improvement. 
First, providing guidance and code-as-policy to \framework FSM-driven \textit{System 1} remains a major challenge for many models with weaker capabilities, especially small models. 
Many models are still limited by errors in their output, which cannot be verified and thus lead to invalid policies.
Secondly, using lambda functions to control the FSM still has a certain lack of flexibility. However, given the current limitations of model capabilities, it might be hard for models to directly output valid state machine code.
And since ToM ability is a complex higher-order reasoning capability, it imposes high demands on the model itself. 
This limitation makes it more likely for ToM failures to occur when \framework is applied to smaller models.
Big models with strong reasoning capabilities suffer from high latency, which reduces the timeliness of reasoning, which is another limitation of \framework, making it challenging to consistently outperform FSM across different models.
Our current experiments are still conducted on a small scale, and since the human subjects are all university students, there may be potential biases. 
Conducting larger-scale experiments in the future will help deepen our understanding of simultaneous human-AI collaboration.

\section*{Acknowledgments}
Team from Shanghai Jiao Tong University is supported by the National Key R\&D Program of China (2024YFC3505402), National Natural Science Foundation of China (T2421002, 62061146002, 62020106005, U2244217, 62322603), Shanghai Municipal Science and Technology Major Project (2021SHZDZX0102), and Shanghai Pilot Program for Basic Research - Shanghai Jiao Tong University.
Xihuai Wang is supported by the Wen-Tsun Wu AI Honorary Doctoral Scholarship from AI Institute, Shanghai Jiao Tong University. 
We extend heartfelt thanks to the participants in our human experiments from Shanghai Jiao Tong University.
We would like to express our gratitude to the AGI-Eval community for their support in hosting the evaluation results of this work.

\bibliography{dpt}

\begin{thebibliography}{52}
\providecommand{\natexlab}[1]{#1}

\bibitem[{Abdin et~al.(2024)Abdin, Aneja, Behl, Bubeck, Eldan, Gunasekar, Harrison, Hewett, Javaheripi, Kauffmann et~al.}]{abdin2024phi4}
Marah Abdin, Jyoti Aneja, Harkirat Behl, S{\'e}bastien Bubeck, Ronen Eldan, Suriya Gunasekar, Michael Harrison, Russell~J Hewett, Mojan Javaheripi, Piero Kauffmann, et~al. 2024.
\newblock Phi-4 technical report.
\newblock \emph{arXiv preprint arXiv:2412.08905}.

\bibitem[{AI()}]{mistralAI}
Mistral AI.
\newblock \href {https://mistral.ai/en/models} {Mistral ai - models}.

\bibitem[{Anthropic(2024)}]{anthropic_claude35}
Anthropic. 2024.
\newblock Claude 3.5 models and computer use.
\newblock \url{https://www.anthropic.com/news/3-5-models-and-computer-use}.

\bibitem[{Ashktorab et~al.(2021)Ashktorab, Dugan, Johnson, Pan, Zhang, Kumaravel, and Campbell}]{zahra2021direction}
Zahra Ashktorab, Casey Dugan, James Johnson, Qian Pan, Wei Zhang, Sadhana Kumaravel, and Murray Campbell. 2021.
\newblock \href {https://doi.org/10.1145/3411764.3445256} {Effects of communication directionality and ai agent differences in human-ai interaction}.
\newblock In \emph{Proceedings of the 2021 CHI Conference on Human Factors in Computing Systems}, CHI '21, New York, NY, USA. Association for Computing Machinery.

\bibitem[{Baron-Cohen et~al.(1985)Baron-Cohen, Leslie, and Frith}]{baron1985does}
Simon Baron-Cohen, Alan~M Leslie, and Uta Frith. 1985.
\newblock Does the autistic child have a “theory of mind”?
\newblock \emph{Cognition}, 21(1):37--46.

\bibitem[{Bernstein et~al.(2002)Bernstein, Givan, Immerman, and Zilberstein}]{bernstein2002complexity}
Daniel~S Bernstein, Robert Givan, Neil Immerman, and Shlomo Zilberstein. 2002.
\newblock The complexity of decentralized control of markov decision processes.
\newblock \emph{Mathematics of operations research}, 27(4):819--840.

\bibitem[{Carroll et~al.(2019)Carroll, Shah, Ho, Griffiths, Seshia, Abbeel, and Dragan}]{carroll2019utility}
Micah Carroll, Rohin Shah, Mark~K Ho, Tom Griffiths, Sanjit Seshia, Pieter Abbeel, and Anca Dragan. 2019.
\newblock On the utility of learning about humans for human-ai coordination.
\newblock \emph{Advances in neural information processing systems}, 32.

\bibitem[{Contributors(2023)}]{ollama}
Ollama Contributors. 2023.
\newblock \href {https://github.com/ollama/ollama} {Ollama: Run large language models locally}.

\bibitem[{Dafoe et~al.(2021)Dafoe, Bachrach, Hadfield, Horvitz, Larson, and Graepel}]{dafoe2021cooperative}
Allan Dafoe, Yoram Bachrach, Gillian Hadfield, Eric Horvitz, Kate Larson, and Thore Graepel. 2021.
\newblock Cooperative ai: machines must learn to find common ground.

\bibitem[{Dourish and Bellotti(1992)}]{collaborationawareness1992}
Paul Dourish and Victoria Bellotti. 1992.
\newblock \href {https://doi.org/10.1145/143457.143468} {Awareness and coordination in shared workspaces}.
\newblock In \emph{Proceedings of the 1992 ACM Conference on Computer-Supported Cooperative Work}, CSCW '92, page 107–114, New York, NY, USA. Association for Computing Machinery.

\bibitem[{Evans and Stanovich(2013)}]{evans2013dual}
Jonathan St~BT Evans and Keith~E Stanovich. 2013.
\newblock Dual-process theories of higher cognition: Advancing the debate.
\newblock \emph{Perspectives on psychological science}, 8(3):223--241.

\bibitem[{Gerganov(2023)}]{llama.cpp}
Georgi Gerganov. 2023.
\newblock \href {https://github.com/ggerganov/llama.cpp} {llama.cpp: Port of meta's llama model in c/c++}.

\bibitem[{Gong et~al.(2024)Gong, Huang, Ma, Noda, Durante, Zheng, Terzopoulos, Fei-Fei, Gao, and Vo}]{gong2024mindagent}
Ran Gong, Qiuyuan Huang, Xiaojian Ma, Yusuke Noda, Zane Durante, Zilong Zheng, Demetri Terzopoulos, Li~Fei-Fei, Jianfeng Gao, and Hoi Vo. 2024.
\newblock Mindagent: Emergent gaming interaction.
\newblock In \emph{Findings of the Association for Computational Linguistics: NAACL 2024}, pages 3154--3183.

\bibitem[{Guan et~al.(2023)Guan, Zhang, Fan, Li, Chen, Li, Tian, Yuan, and Yu}]{guan2023efficient}
Cong Guan, Lichao Zhang, Chunpeng Fan, Yichen Li, Feng Chen, Lihe Li, Yunjia Tian, Lei Yuan, and Yang Yu. 2023.
\newblock Efficient human-ai coordination via preparatory language-based convention.
\newblock \emph{arXiv preprint arXiv:2311.00416}.

\bibitem[{Guo et~al.(2025)Guo, Yang, Zhang, Song, Zhang, Xu, Zhu, Ma, Wang, Bi et~al.}]{guo2025deepseek}
Daya Guo, Dejian Yang, Haowei Zhang, Junxiao Song, Ruoyu Zhang, Runxin Xu, Qihao Zhu, Shirong Ma, Peiyi Wang, Xiao Bi, et~al. 2025.
\newblock Deepseek-r1: Incentivizing reasoning capability in llms via reinforcement learning.
\newblock \emph{arXiv preprint arXiv:2501.12948}.

\bibitem[{Hart et~al.(1968)Hart, Nilsson, and Raphael}]{hart1968formal}
Peter~E Hart, Nils~J Nilsson, and Bertram Raphael. 1968.
\newblock A formal basis for the heuristic determination of minimum cost paths.
\newblock \emph{IEEE transactions on Systems Science and Cybernetics}, 4(2):100--107.

\bibitem[{He et~al.(2024)He, Liao, Cao, Liu, Liu, Chen, and Qin}]{he-etal-2024-planning}
Tao He, Lizi Liao, Yixin Cao, Yuanxing Liu, Ming Liu, Zerui Chen, and Bing Qin. 2024.
\newblock \href {https://doi.org/10.18653/v1/2024.acl-long.262} {Planning like human: A dual-process framework for dialogue planning}.
\newblock In \emph{Proceedings of the 62nd Annual Meeting of the Association for Computational Linguistics (Volume 1: Long Papers)}, pages 4768--4791, Bangkok, Thailand. Association for Computational Linguistics.

\bibitem[{Kahneman(2011)}]{kahneman2011thinking}
Daniel Kahneman. 2011.
\newblock Thinking, fast and slow.
\newblock \emph{Farrar, Straus and Giroux}.

\bibitem[{Krych-Appelbaum et~al.(2007)Krych-Appelbaum, Law, Jones, Barnacz, Johnson, and Keenan}]{krych2007think}
Meredyth Krych-Appelbaum, Julie~Banzon Law, Dayna Jones, Allyson Barnacz, Amanda Johnson, and Julian~Paul Keenan. 2007.
\newblock “i think i know what you mean”: The role of theory of mind in collaborative communication.
\newblock \emph{Interaction Studies}, 8(2):267--280.

\bibitem[{Kwon et~al.(2023)Kwon, Li, Zhuang, Sheng, Zheng, Yu, Gonzalez, Zhang, and Stoica}]{kwon2023efficient}
Woosuk Kwon, Zhuohan Li, Siyuan Zhuang, Ying Sheng, Lianmin Zheng, Cody~Hao Yu, Joseph~E. Gonzalez, Hao Zhang, and Ion Stoica. 2023.
\newblock Efficient memory management for large language model serving with pagedattention.
\newblock In \emph{Proceedings of the ACM SIGOPS 29th Symposium on Operating Systems Principles}.

\bibitem[{Li et~al.(2023)Li, Zhang, Sun, Du, Wen, Wang, and Pan}]{Yang23Cole}
Yang Li, Shao Zhang, Jichen Sun, Yali Du, Ying Wen, Xinbing Wang, and Wei Pan. 2023.
\newblock Cooperative open-ended learning framework for zero-shot coordination.
\newblock In \emph{{ICML}}, volume 202 of \emph{Proceedings of Machine Learning Research}, pages 20470--20484. {PMLR}.

\bibitem[{Li et~al.(2024)Li, Zhang, Sun, Zhang, Du, Wen, Wang, and Pan}]{li2024tackling}
Yang Li, Shao Zhang, Jichen Sun, Wenhao Zhang, Yali Du, Ying Wen, Xinbing Wang, and Wei Pan. 2024.
\newblock Tackling cooperative incompatibility for zero-shot human-ai coordination.
\newblock \emph{Journal of Artificial Intelligence Research}, 80:1139--1185.

\bibitem[{Liang et~al.(2023)Liang, Huang, Xia, Xu, Hausman, Ichter, Florence, and Zeng}]{DBLP:conf/icra/LiangHXXHIFZ23}
Jacky Liang, Wenlong Huang, Fei Xia, Peng Xu, Karol Hausman, Brian Ichter, Pete Florence, and Andy Zeng. 2023.
\newblock \href {https://doi.org/10.1109/ICRA48891.2023.10160591} {Code as policies: Language model programs for embodied control}.
\newblock In \emph{{IEEE} International Conference on Robotics and Automation, {ICRA} 2023, London, UK, May 29 - June 2, 2023}, pages 9493--9500. {IEEE}.

\bibitem[{Liu et~al.(2024{\natexlab{a}})Liu, Feng, Xue, Wang, Wu, Lu, Zhao, Deng, Zhang, Ruan et~al.}]{liu2024deepseek}
Aixin Liu, Bei Feng, Bing Xue, Bingxuan Wang, Bochao Wu, Chengda Lu, Chenggang Zhao, Chengqi Deng, Chenyu Zhang, Chong Ruan, et~al. 2024{\natexlab{a}}.
\newblock Deepseek-v3 technical report.
\newblock \emph{arXiv preprint arXiv:2412.19437}.

\bibitem[{Liu et~al.(2024{\natexlab{b}})Liu, Yu, Gao, Xie, Liao, Wu, and Wang}]{liu2024slow}
Jijia Liu, Chao Yu, Jiaxuan Gao, Yuqing Xie, Qingmin Liao, Yi~Wu, and Yu~Wang. 2024{\natexlab{b}}.
\newblock Llm-powered hierarchical language agent for real-time human-ai coordination.
\newblock In \emph{Proceedings of the 23rd International Conference on Autonomous Agents and Multiagent Systems}, AAMAS '24, page 1219–1228, Richland, SC. International Foundation for Autonomous Agents and Multiagent Systems.

\bibitem[{OpenAI(2024)}]{openai2024gpt4omini}
OpenAI. 2024.
\newblock Gpt-4o mini: Advancing cost-efficient intelligence.
\newblock Accessed: 2024-09-05.

\bibitem[{Prather et~al.(2024)Prather, Reeves, Leinonen, MacNeil, Randrianasolo, Becker, Kimmel, Wright, and Briggs}]{prather2024widening}
James Prather, Brent~N Reeves, Juho Leinonen, Stephen MacNeil, Arisoa~S Randrianasolo, Brett~A Becker, Bailey Kimmel, Jared Wright, and Ben Briggs. 2024.
\newblock The widening gap: The benefits and harms of generative ai for novice programmers.
\newblock In \emph{Proceedings of the 2024 ACM Conference on International Computing Education Research-Volume 1}, pages 469--486.

\bibitem[{Premack and Woodruff(1978)}]{premack1978does}
David Premack and Guy Woodruff. 1978.
\newblock Does the chimpanzee have a theory of mind?
\newblock \emph{Behavioral and brain sciences}, 1(4):515--526.

\bibitem[{Rabinowitz et~al.(2018)Rabinowitz, Perbet, Song, Zhang, Eslami, and Botvinick}]{neil2018tom}
Neil Rabinowitz, Frank Perbet, Francis Song, Chiyuan Zhang, S.~M.~Ali Eslami, and Matthew Botvinick. 2018.
\newblock \href {https://proceedings.mlr.press/v80/rabinowitz18a.html} {Machine theory of mind}.
\newblock In \emph{Proceedings of the 35th International Conference on Machine Learning}, volume~80 of \emph{Proceedings of Machine Learning Research}, pages 4218--4227. PMLR.

\bibitem[{Riemer et~al.(2024)Riemer, Ashktorab, Bouneffouf, Das, Liu, Weisz, and Campbell}]{riemer2024can}
Matthew Riemer, Zahra Ashktorab, Djallel Bouneffouf, Payel Das, Miao Liu, Justin~D Weisz, and Murray Campbell. 2024.
\newblock Can large language models adapt to other agents in-context?
\newblock \emph{arXiv preprint arXiv:2412.19726}.

\bibitem[{Russell and Norvig(2016)}]{russell2016artificial}
Stuart~J Russell and Peter Norvig. 2016.
\newblock \emph{Artificial intelligence: a modern approach}.
\newblock Pearson.

\bibitem[{Salikutluk et~al.(2024)Salikutluk, Sch\"{o}pper, Herbert, Scheuermann, Frodl, Balfanz, J\"{a}kel, and Koert}]{vildan2024auto}
Vildan Salikutluk, Janik Sch\"{o}pper, Franziska Herbert, Katrin Scheuermann, Eric Frodl, Dirk Balfanz, Frank J\"{a}kel, and Dorothea Koert. 2024.
\newblock \href {https://doi.org/10.1145/3613904.3642564} {An evaluation of situational autonomy for human-ai collaboration in a shared workspace setting}.
\newblock In \emph{Proceedings of the CHI Conference on Human Factors in Computing Systems}, CHI '24, New York, NY, USA. Association for Computing Machinery.

\bibitem[{Shao et~al.(2024)Shao, Samuel, Jiang, Yang, and Yang}]{shao2024collaborative}
Yijia Shao, Vinay Samuel, Yucheng Jiang, John Yang, and Diyi Yang. 2024.
\newblock Collaborative gym: A framework for enabling and evaluating human-agent collaboration.
\newblock \emph{arXiv preprint arXiv:2412.15701}.

\bibitem[{Shinn et~al.(2024)Shinn, Cassano, Gopinath, Narasimhan, and Yao}]{shinn2024reflexion}
Noah Shinn, Federico Cassano, Ashwin Gopinath, Karthik Narasimhan, and Shunyu Yao. 2024.
\newblock Reflexion: Language agents with verbal reinforcement learning.
\newblock \emph{Advances in Neural Information Processing Systems}, 36.

\bibitem[{Strouse et~al.(2021)Strouse, McKee, Botvinick, Hughes, and Everett}]{strouse2021fcp}
DJ~Strouse, Kevin McKee, Matt Botvinick, Edward Hughes, and Richard Everett. 2021.
\newblock Collaborating with humans without human data.
\newblock \emph{Advances in Neural Information Processing Systems}, 34:14502--14515.

\bibitem[{Team et~al.(2024)Team, Riviere, Pathak, Sessa, Hardin, Bhupatiraju, Hussenot, Mesnard, Shahriari, Ram{\'e} et~al.}]{team2024gemma}
Gemma Team, Morgane Riviere, Shreya Pathak, Pier~Giuseppe Sessa, Cassidy Hardin, Surya Bhupatiraju, L{\'e}onard Hussenot, Thomas Mesnard, Bobak Shahriari, Alexandre Ram{\'e}, et~al. 2024.
\newblock Gemma 2: Improving open language models at a practical size.
\newblock \emph{arXiv preprint arXiv:2408.00118}.

\bibitem[{Touvron et~al.(2023)Touvron, Lavril, Izacard, Martinet, Lachaux, Lacroix, Rozi{\`e}re, Goyal, Hambro, Azhar et~al.}]{touvron2023llama}
Hugo Touvron, Thibaut Lavril, Gautier Izacard, Xavier Martinet, Marie-Anne Lachaux, Timoth{\'e}e Lacroix, Baptiste Rozi{\`e}re, Naman Goyal, Eric Hambro, Faisal Azhar, et~al. 2023.
\newblock Llama: Open and efficient foundation language models.
\newblock \emph{arXiv preprint arXiv:2302.13971}.

\bibitem[{Wan et~al.(2024)Wan, Hu, Zhang, Wang, Wen, and Lu}]{wan2024felt}
Qian Wan, Siying Hu, Yu~Zhang, Piaohong Wang, Bo~Wen, and Zhicong Lu. 2024.
\newblock " it felt like having a second mind": Investigating human-ai co-creativity in prewriting with large language models.
\newblock \emph{Proceedings of the ACM on Human-Computer Interaction}, 8(CSCW1):1--26.

\bibitem[{Wang et~al.(2024)Wang, Zhang, Zhang, Dong, Chen, Wen, and Zhang}]{wang2024zsc}
Xihuai Wang, Shao Zhang, Wenhao Zhang, Wentao Dong, Jingxiao Chen, Ying Wen, and Weinan Zhang. 2024.
\newblock Zsc-eval: An evaluation toolkit and benchmark for multi-agent zero-shot coordination.
\newblock In \emph{The Thirty-eight Conference on Neural Information Processing Systems Datasets and Benchmarks Track}.

\bibitem[{Wei et~al.(2022)Wei, Wang, Schuurmans, Bosma, Xia, Chi, Le, Zhou et~al.}]{wei2022chain}
Jason Wei, Xuezhi Wang, Dale Schuurmans, Maarten Bosma, Fei Xia, Ed~Chi, Quoc~V Le, Denny Zhou, et~al. 2022.
\newblock Chain-of-thought prompting elicits reasoning in large language models.
\newblock \emph{Advances in neural information processing systems}, 35:24824--24837.

\bibitem[{Wen et~al.(2019)Wen, Yang, Luo, Wang, and Pan}]{wen2018probabilistic}
Ying Wen, Yaodong Yang, Rui Luo, Jun Wang, and Wei Pan. 2019.
\newblock \href {https://openreview.net/forum?id=rkl6As0cF7} {Probabilistic recursive reasoning for multi-agent reinforcement learning}.
\newblock In \emph{International Conference on Learning Representations}.

\bibitem[{Wester et~al.(2024)Wester, Jacobsen, de~Jong, Als, Djern{\ae}s, and van Berkel}]{wester2024theory}
Joel Wester, Rune~M{\o}berg Jacobsen, Sander de~Jong, Naja Kathrine~Kollerup Als, Helena~B{\o}jer Djern{\ae}s, and Niels van Berkel. 2024.
\newblock Theory of mind and self-presentation in human-llm interactions.
\newblock In \emph{Adjunct Proceedings of the ACM SIGCHI Conference on Human Factors in Computing Systems}.

\bibitem[{Wu et~al.(2021)Wu, Wang, Evans, Tenenbaum, Parkes, and Kleiman-Weiner}]{gymcooking}
Sarah~A. Wu, Rose~E. Wang, James~A. Evans, Joshua~B. Tenenbaum, David~C. Parkes, and Max Kleiman-Weiner. 2021.
\newblock \href {https://doi.org/10.1111/tops.12525} {Too many cooks: Bayesian inference for coordinating multi-agent collaboration}.
\newblock \emph{Topics in Cognitive Science}, 13(2):414--432.

\bibitem[{Yang et~al.(2024)Yang, Yang, Zhang, Hui, Zheng, Yu, Li, Liu, Huang, Wei, Lin, Yang, Tu, Zhang, Yang, Yang, Zhou, Lin, Dang, Lu, Bao, Yang, Yu, Li, Xue, Zhang, Zhu, Men, Lin, Li, Xia, Ren, Ren, Fan, Su, Zhang, Wan, Liu, Cui, Zhang, and Qiu}]{qwen2.5}
An~Yang, Baosong Yang, Beichen Zhang, Binyuan Hui, Bo~Zheng, Bowen Yu, Chengyuan Li, Dayiheng Liu, Fei Huang, Haoran Wei, Huan Lin, Jian Yang, Jianhong Tu, Jianwei Zhang, Jianxin Yang, Jiaxi Yang, Jingren Zhou, Junyang Lin, Kai Dang, Keming Lu, Keqin Bao, Kexin Yang, Le~Yu, Mei Li, Mingfeng Xue, Pei Zhang, Qin Zhu, Rui Men, Runji Lin, Tianhao Li, Tingyu Xia, Xingzhang Ren, Xuancheng Ren, Yang Fan, Yang Su, Yichang Zhang, Yu~Wan, Yuqiong Liu, Zeyu Cui, Zhenru Zhang, and Zihan Qiu. 2024.
\newblock Qwen2.5 technical report.
\newblock \emph{arXiv preprint arXiv:2412.15115}.

\bibitem[{Yao et~al.(2022)Yao, Zhao, Yu, Du, Shafran, Narasimhan, and Cao}]{yao2022react}
Shunyu Yao, Jeffrey Zhao, Dian Yu, Nan Du, Izhak Shafran, Karthik Narasimhan, and Yuan Cao. 2022.
\newblock React: Synergizing reasoning and acting in language models.
\newblock \emph{arXiv preprint arXiv:2210.03629}.

\bibitem[{Yi et~al.(2024)Yi, Ouyang, Liu, Liao, Xu, and Shen}]{yi2024survey}
Zihao Yi, Jiarui Ouyang, Yuwen Liu, Tianhao Liao, Zhe Xu, and Ying Shen. 2024.
\newblock A survey on recent advances in llm-based multi-turn dialogue systems.
\newblock \emph{arXiv preprint arXiv:2402.18013}.

\bibitem[{Yu et~al.(2023)Yu, Gao, Liu, Xu, Tang, Yang, Wang, and Wu}]{yu23hsp}
Chao Yu, Jiaxuan Gao, Weilin Liu, Botian Xu, Hao Tang, Jiaqi Yang, Yu~Wang, and Yi~Wu. 2023.
\newblock Learning zero-shot cooperation with humans, assuming humans are biased.
\newblock In \emph{{ICLR}}. OpenReview.net.

\bibitem[{Yu et~al.(2024)Yu, Xu, Weston, and Kulikov}]{yu2024distilling}
Ping Yu, Jing Xu, Jason Weston, and Ilia Kulikov. 2024.
\newblock Distilling system 2 into system 1.
\newblock \emph{arXiv preprint arXiv:2407.06023}.

\bibitem[{Zhang et~al.(2024{\natexlab{a}})Zhang, Yang, Hu, Wang, Li, Sun, Zhang, Zhang, Liu, Zhu et~al.}]{zhang2024proagent}
Ceyao Zhang, Kaijie Yang, Siyi Hu, Zihao Wang, Guanghe Li, Yihang Sun, Cheng Zhang, Zhaowei Zhang, Anji Liu, Song-Chun Zhu, et~al. 2024{\natexlab{a}}.
\newblock Proagent: building proactive cooperative agents with large language models.
\newblock In \emph{Proceedings of the AAAI Conference on Artificial Intelligence}, volume~38, pages 17591--17599.

\bibitem[{Zhang et~al.(2024{\natexlab{b}})Zhang, Wang, Zhang, Chen, Gao, Wang, Zhang, Wang, and Wen}]{zhang2024mutual}
Shao Zhang, Xihuai Wang, Wenhao Zhang, Yongshan Chen, Landi Gao, Dakuo Wang, Weinan Zhang, Xinbing Wang, and Ying Wen. 2024{\natexlab{b}}.
\newblock Mutual theory of mind in human-ai collaboration: An empirical study with llm-driven ai agents in a real-time shared workspace task.
\newblock \emph{arXiv preprint arXiv:2409.08811}.

\bibitem[{Zhang et~al.(2024{\natexlab{c}})Zhang, Tang, Wu, Wang, Shen, Hou, Tan, Li, Zhuang, and Lu}]{DBLP:conf/acl/ZhangTWW0HTLZ024}
Wenqi Zhang, Ke~Tang, Hai Wu, Mengna Wang, Yongliang Shen, Guiyang Hou, Zeqi Tan, Peng Li, Yueting Zhuang, and Weiming Lu. 2024{\natexlab{c}}.
\newblock \href {https://aclanthology.org/2024.acl-long.292} {Agent-pro: Learning to evolve via policy-level reflection and optimization}.
\newblock In \emph{Proceedings of the 62nd Annual Meeting of the Association for Computational Linguistics (Volume 1: Long Papers), {ACL} 2024, Bangkok, Thailand, August 11-16, 2024}, pages 5348--5375. Association for Computational Linguistics.

\bibitem[{Zhou et~al.(2024)Zhou, Ning, Hong, Fu, Xu, Li, Lou, Wang, Yuan, Li et~al.}]{zhou2024survey}
Zixuan Zhou, Xuefei Ning, Ke~Hong, Tianyu Fu, Jiaming Xu, Shiyao Li, Yuming Lou, Luning Wang, Zhihang Yuan, Xiuhong Li, et~al. 2024.
\newblock A survey on efficient inference for large language models.
\newblock \emph{arXiv preprint arXiv:2404.14294}.

\end{thebibliography}

\clearpage
\appendix

\section{Environment Details}\label{app:env}

We implement the environment from \cite{zhang2024mutual} based on overcooked-ai\footnote{\url{https://github.com/HumanCompatibleAI/overcooked_ai}, MIT License} \cite{carroll2019utility} and gym-cooking\footnote{\url{https://github.com/rosewang2008/gym-cooking}, MIT License} \cite{gymcooking}.

\paragraph{State.} Both the agent and the human have full access to the game states and each other's actions.
Players can directly see the status of all items in the game interface, such as the location where items are placed and their current state (e.g., beef cooking in a pan).
Players can also view the remaining game time and current score through the information displayed. 
The remaining time for each order, the progress of chopping lettuce, the process of cooking beef, and the process of extinguishing a fire are shown through progress bars.
All actions taken by teammates, the teammates' location, and the items they are holding are fully visible to each other.

\begin{figure}[h]
    \centering
    \subfigure[LettuceBurger]{
    \includegraphics[width=0.4\linewidth]{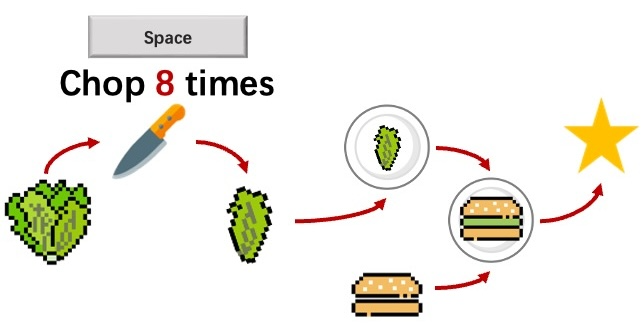}
    \label{fig:vegan}
    }
    % \hfill
    \subfigure[BeefBurger]{
    \includegraphics[width=0.4\linewidth]{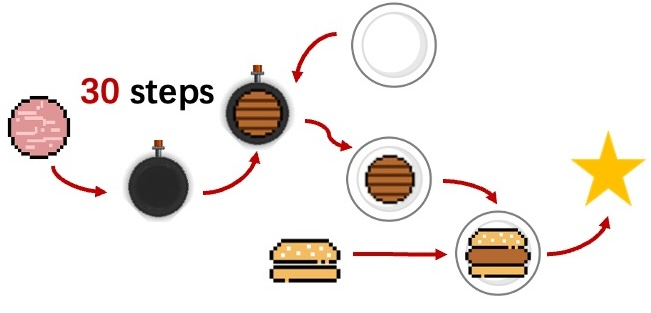}
    \label{fig:meat}
    }
    \hfill
    \subfigure[BeefLettuceBurger]{
    \includegraphics[width=0.4\linewidth]{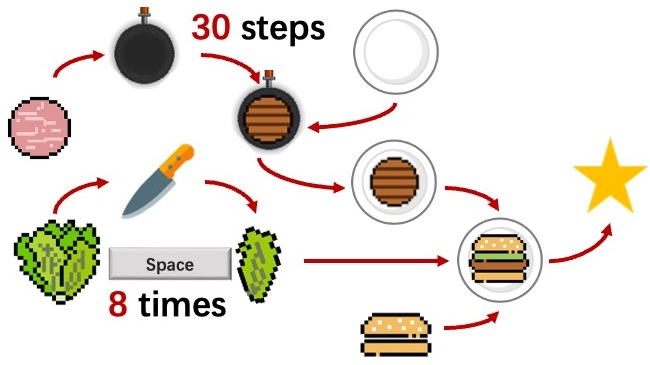}
    \label{fig:allinone}
    }
    % \hfill
    \subfigure[Overcooked Beef]{
    \includegraphics[width=0.4\linewidth]{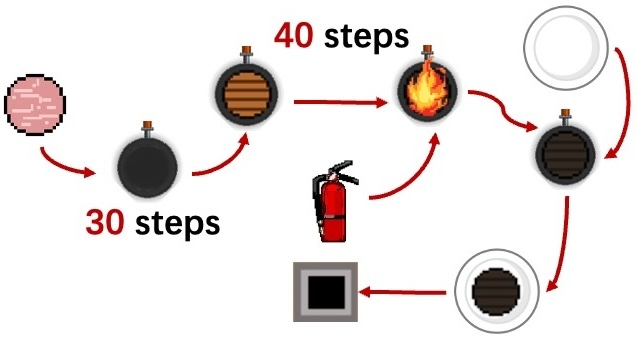}
    \label{fig:fire}
    }
    \caption{\textbf{Game Mechanism from \citet{zhang2024mutual}.} (a), (b), and (c) are the rules for preparing and serving burgers. (d) demonstrates the mechanism of overcooked beef and the rules for handling the fire caused by overcooked beef.}
    \label{fig:diff_level}
\end{figure}

\paragraph{Action.} In this environment, the actions that the human and the agent can take to control the chefs include moving up, down, left, and right, as well as ``interact''. 
All activities such as picking up items, serving dishes, and extinguishing fires are considered as ``interact'' actions.
The specific interaction rules are illustrated in Figure \ref{fig:diff_level}. 
We denote the actions to control the chefs as $\mathcal{A}^{\text{control}}$.
The agent and the human share the same $\mathcal{A}^{\text{control}}$.

\paragraph{Reward.} The scores for completing the three different types of orders vary and serving the wrong burger or missing an order will result in a penalty. The specific rewards are detailed in \Cref{tab:reward}.

\paragraph{Timesteps.} In the environment implementation, one timestep is 0.25 sencond in the real world. At most one action can be executed at each time step.

\begin{table}[]
    \centering

        \resizebox{0.9\linewidth}{!}{
    \begin{tabular}{l|c}
    \toprule
        \textbf{Event} & \textbf{Rewards} \\
        \midrule
        Serve a LettuceBurger & +15\\
        Serve a BeefBurger & +20\\
        Serve a BeefLettuceBurger & +25\\
        Serve a Wrong Burger (or Something not a Burger) & -10\\
        Miss an order & -10\\
\bottomrule
    \end{tabular}
}
        \caption{\textbf{Rewards in Game.}}
    \label{tab:reward}
\end{table}

\subsection{FSM in Overcooked}
\label{app:fsm}

Follow the \citet{zhang2024mutual}, the macro actions included are summarized below.
\begin{itemize}
    \item \textbf{Prepare:}  
    \begin{itemize}
        \item \textbf{Valid Objects:} ``Beef'', ``Lettuce'', ``Bread''
        \item \textbf{Function:} Prepare an appointed ingredient until it can be used to assemble.
    \end{itemize}
    
    \item \textbf{Assemble:}  
    \begin{itemize}
        \item \textbf{Valid Objects:} ``BeefBurger'', ``LettuceBurger'', ``BeefLettuceBurger''
        \item \textbf{Function:} Assemble an appointed burger if all necessary ingredients are ready.
    \end{itemize}
    
    \item \textbf{Pass on:}  
    \begin{itemize}
        \item \textbf{Valid Objects:} ``Plate'', ``Bread''
        \item \textbf{Function:} Put the object onto the center counters to deliver it to the partner.
    \end{itemize}
    
    \item \textbf{Serve:}  
    \begin{itemize}
        \item \textbf{Valid Objects:} ``BeefBurger'', ``LettuceBurger'', ``BeefLettuceBurger''
        \item \textbf{Function:} Deliver an assembled burger to the customer.
    \end{itemize}
    
    \item \textbf{Putout Fire:}  
    \begin{itemize}
        \item \textbf{Valid Objects:} -
        \item \textbf{Function:} Pick up the fire extinguisher and put out the fire, if any.
    \end{itemize}
\end{itemize}

\onecolumn
\section{General Game Prompt of All Experiments}\label{app:gameprompt}

\begin{lstlisting}

-------------------------------System Prompt-------------------------------

As a player in a collaborative cooking game, you are working with a human player to complete hamburger orders.
Focus on cooperation, player engagement, fulfillment, and score accrual.

-------------------------------Game Prompt-------------------------------

# Game Introduction

## Game Scene

The game environment is set in a kitchen, designed for a collaborative cooking challenge. The layout includes a central counter area surrounded by various stations and essential elements for gameplay. Here's a detailed breakdown of the scene:

- **Central Counter Area**: The central space has a counter where ingredients can be placed temporarily for efficient workflow.
- **Ingredient Stations**: Distribution stations for picking up `Lettuce`, `Beef` and `Bread`.
- **Cooking and Preparation Tools**:
  - **Pans**: for cooking `Beef`.
  - **Cutboards**: for preparing `Lettuce`.
- **Plate Station**: for picking up empty plates.
- **Fire Extinguisher**: for extinguishing fires and can be moved.
- **Serving Area**: for serving orders.

You are controlling one of the two chefs in the kitchen, and your goal is to work together with your partner to fulfill customer orders efficiently and accurately by writing codes to improve your policies in the game.

## Game Mechanisms

### Game Objects

Each object is a represented as a tuple of `(object_name, object_status)`.
    - **Beef**: Includes `("Beef", "Fresh")`, `("Beef", "In-progress")`, `("Beef", "Well-cooked")`, `("Beef", "Overcooked")`. Note that `("Beef", "In-progress")` will become `("Beef", "Well-cooked")` after a certain time, and `("Beef", "Well-cooked")` will become `("Beef", "Overcooked")` if left on the pan for too long.
    - **Lettuce**: Includes `("Lettuce", "Unchopped")` and `("Lettuce", "Chopped")`.
    - **Bread**: Represented as `("Bread", "")`.
    - **BeefLettuce**: A mixture of ingredients, represented as `("BeefLettuce", "")`.
    - **Burgers**: Types include `("BeefBurger", "")`, `("LettuceBurger", "")`, and `("BeefLettuceBurger", "")`.
    - **Plate**: Represented as `("Plate", "")`.
    - **FireExtinguisher**: Represented as `("FireExtinguisher", "")`.
    - **Fire**: Indicates an active fire, represented as `("Fire", "")`.

### Counters

Especially, we count the status of the counters in the kitchen:
    - "Empty": No object on the counter.

### Valid Actions in Code

To play the game, you can use the following actions:

- **Prepare Actions**: Used to prepare individual ingredients. Each ingredient can be prepared with the option to either place it on a plate or not. Here are the valid prepare actions:
    - Preparing `("Beef", "Well-cooked")`: Get a `("Beef", "Fresh")` and cook it into a `("Beef", "In-progress")` and then a `("Beef", "Well-cooked")`. Pay attention to avoid overcooking it, which will result in a `("Beef", "Overcooked")` and a `("Fire", "")` in the pan.
      - `("prepare", {"food": "Beef", "plate": True})`
      - `("prepare", {"food": "Beef", "plate": False})`
    - Preparing `("Lettuce", "Chopped")`: Get a `("Lettuce", "Unchopped")` and chop it into a `("Lettuce", "Chopped")`.
        - `("prepare", {"food": "Lettuce", "plate": True})`
        - `("prepare", {"food": "Lettuce", "plate": False})`
    - Preparing `("Bread", "")`: Get a `("Bread", "")` and put it on the counter or in a plate.
      - `("prepare", {"food": "Bread", "plate": True})`
      - `("prepare", {"food": "Bread", "plate": False})`

    Note that when there is no `("Beef", "Fresh")`, `("Lettuce", "Unchopped")` or `("Bread", "")` in the kitchen, the prepare actions will automatically get the ingredients from the respective stations.

- **Assemble Actions**: Used to assemble burgers with the already prepared ingredients (`("Beef", "Well-cooked")`, `("Lettuce", "Chopped")` and `("Bread", "")`). These actions will only be performed if all required ingredients are ready. See the **Cookbook** section for the burger types and their ingredients.
  - `("assemble", {"food": "LettuceBurger"})`
  - `("assemble", {"food": "BeefBurger"})`
  - `("assemble", {"food": "BeefLettuceBurger"})`

- **Pass On Action**: Used to pass something to your partner by putting it on the central counter.
  - `("pass_on", {"thing": "Plate"})`
  - `("pass_on", {"thing": "Bread"})`
  - `("pass_on", {"thing": "Lettuce", "thing_status": "Chopped"})`
  - `("pass_on", {"thing": "Lettuce", "thing_status": "Unchopped"})`
  - `("pass_on", {"thing": "Beef", "thing_status": "Well-cooked"})`
  - `("pass_on", {"thing": "Beef", "thing_status": "Fresh"})`
  - `("pass_on", {"thing": "BeefLettuce"})`
  - `("pass_on", {"thing": "BeefBurger"})`
  - `("pass_on", {"thing": "LettuceBurger"})`
  - `("pass_on", {"thing": "BeefLettuceBurger"})`
  - `("pass_on", {"thing": "FireExtinguisher"})`

- **Serve Actions**: Used to serve the assembled burgers to the customer.
  - `("serve", {"food": "BeefBurger"})`
  - `("serve", {"food": "LettuceBurger"})`
  - `("serve", {"food": "BeefLettuceBurger"})`

- **Put Out Fire Action**: Used to pick up the fire extinguisher and put out the fire on the pan when the `Beef` is overcooked and catches fire.
  - `("putout_fire", {})`

- **Clean A Counter Action**: Used to clean a counter by dropping all objects on it to the trash can.
  - `("clean_a_counter", {})`

### Cookbook

In this collaborative kitchen game, the goal is to prepare and serve burgers efficiently to earn points. The game features three types of burgers: `LettuceBurger`, `BeefBurger`, and `BeefLettuceBurger`. Here are the rules and how the actions fit into the gameplay:

- **LettuceBurger**:
  - **Ingredients**: `("Lettuce", "Chopped")`, `("Bread", "")`
  - **Preparation**:
    - Prepare `("Lettuce", "Chopped")` if not already prepared
    - Assemble the ingredients using the action: `("assemble", {"food": "LettuceBurger"})`
- **BeefBurger**:
  - **Ingredients**: `("Beef", "Well-cooked")`, `("Bread", "")`
  - **Preparation**:
    - Prepare `("Beef", "Well-cooked")`  if not already prepared
    - Assemble the ingredients using the action: `("assemble", {"food": "BeefBurger"})`
- **BeefLettuceBurger**:
  - **Ingredients**: `("Lettuce", "Chopped")`, `("Beef", "Well-cooked")`, `("Bread", "")`
  - **Preparation**:
    - Method One
        - Prepare `("Lettuce", "Chopped")` if not already prepared
        - Prepare `("Beef", "Well-cooked")` if not already prepared
        - Assemble the ingredients using the action: `("assemble", {"food": "BeefLettuceBurger"})`
    - Method Two
        - Prepare `("Lettuce", "Chopped")` if not already prepared
        - Assemble the ingredients using the action: `("assemble", {"food": "LettuceBurger"})`
        - Prepare `("Beef", "Well-cooked")` if not already prepared
        - Assemble the ingredients using the action: `("assemble", {"food": "BeefLettuceBurger"})`
    - Method Three
        - Prepare `("Beef", "Well-cooked")` if not already prepared
        - Assemble the ingredients using the action: `("assemble", {"food": "BeefBurger"})`
        - Prepare `("Lettuce", "Chopped")` if not already prepared
        - Assemble the ingredients using the action: `("assemble", {"food": "BeefLettuceBurger"})`
    - Method Four
        - If there is a prepared `("BeefLettuce", "")`, you can directly assemble the `BeefLettuceBurger` using the action: `("assemble", {"food": "BeefLettuceBurger"})`
Note:
- The `Bread` will be automatically used, from prepared `Bread` or the Bread Station, when the `assemble` action is performed. You can also prepare `Bread` in advance.
- **Preparation Flexibility**: You can complete a burger in a flexible order. For example, when making a `BeefLettuceBurger', you can prepare the `Lettuce` before the `Beef`, or vice versa.


### Scoring System

- **Points Earned**:
  There are orders from customers that need to be fulfilled. Each order has a specific point value:
  - `LettuceBurger`: 15 points
  - `BeefBurger`: 20 points
  - `BeefLettuceBurger`: 25 points
- **Points Lost**:
  - Missing an order results in losing 10 points. Ensure that each order is completed within the given time.
  - Serving an item that is not in the order lists also results in losing 10 points. Make sure only demanded burgers are served to customers.


### Important Tips
- **Unreachable Orders**: If the remaining time for an order is less than the time required to prepare the ingredients, it is better to skip that order and focus on the next one.
\end{lstlisting}

\section{\framework Implementation in Overcooked}\label{app:imple}

\onecolumn

\subsection{Instruction Prompt}\label{app:instruct}

\subsubsection{Game State Example}
\begin{lstlisting}[language=Python]
{
    "objects": {
        ("Beef", "Fresh"): 1,
        ("Beef", "In-progress"): 1,
        ("Beef", "Well-cooked"): 0,
        ("Beef", "Overcooked"): 1,
        ("Lettuce", "Unchopped"): 3,
        ("Lettuce", "Chopped"): 1,
        ("Bread", ""): 4,
        ("BeefLettuce", ""): 0,
        ("BeefBurger", ""): 0,
        ("LettuceBurger", ""): 1,
        ("BeefLettuceBurger", ""): 0,
        ("Plate", "Empty"): 2,
        ("FireExtinguisher", ""): 1,
        ("Fire", ""): 0
    },
    "counters": {
        "Empty": 18,
    },
    "orders": [
        {
            "name": "BeefBurger",
            "remain_time": 30
        },
        {
            "name": "LettuceBurger",
            "remain_time": 45
        }
    ],
    "inventory_other_player": {
        "player_1": ("Plate", "Empty"),
    }
}
\end{lstlisting}

\subsubsection{Assigned Task Example}
\begin{lstlisting}[language=Python]
[
    (
        "lambda json_state: json_state['objects'][('Beef', 'Well-cooked')] + json_state['objects'][('Beef', 'In-progress')] < sum(order['name'] == 'BeefBurger' or order['name'] == 'BeefLettuceBurger' for order in json_state['orders'])", ("prepare", {"food": "Beef", "plate": False})
    ),
    "BeefBurger",
    "LettuceBurger"
]
\end{lstlisting}

\subsubsection{Instructions}
\begin{lstlisting}
# Instructions

## Goal

Based on these settings, you need to consider how to play the game with your partner to achieve a higher score. The agent will automatically prepare the burger order with the least remaining time. You will receive game history and your task is to respond to urgent situations for improving the performance.

## Input Information

**Game History**:
    - A sequence of game scenes that have occurred in the past. Each game scene is consisted of:
        - Remained Timestep: The remained timestep of the game.
        - Score: The current score of the game.
        - Game State: The occurrences of objects, orders, and other players' inventories.
        - Action: Actions taken by your agent and the human-controlled agent.
        - Delivery: The food that have been delivered and the corresponding obtained score.
        - Missed Orders: The orders that have not been completed in time and the obtained punished score.
        {MESSAGE_PROMPT}
**Current Assigned Tasks**:
    - The current actions and orders you assigned to the agent that need to be done urgently.
**Behavior Guidelines**:
    - The behavior guidelines are the suggestions you have given to the agent based on the game history.
{INFERRED_HUMAN_PROMPT}

### Game State

The current state of the game includes various details. Here's a detailed description based on the provided structure:

1. **Objects**:
    - The `objects` dictionary records the number of objects with different statuses. Each entry is a tuple of `(object_name, object_status)` mapped to `object_number`.
    - For example:
        - `("LettuceBurger", "") : 1` indicates that there is 1 `LettuceBurger`.

2. **Orders**:
    - The `orders` list contains the current orders that need to be completed. Each order is a dictionary with:
        - `name`: The name of the order, which can be `BeefBurger`, `LettuceBurger`, or `BeefLettuceBurger`.
        - `remain_time`: The remaining time to complete the order, with smaller remaining time indicating higher urgency.

3. **Inventory of the Other Player**:
    - The `inventory_other_player` dictionary records the objects held by the other player. Each entry maps `other_agent_id` to a tuple of `(object_name, object_status)`.
    - This helps in understanding what the other player is currently holding, allowing for better coordination.

#### Example Game State

Now I will show you a game state example. In this example, there are
- 1 fresh beef, 1 in-progress beef, and 1 overcooked beef. No well-cooked beef.
- 3 unchopped lettuce and 1 chopped lettuce.
- 4 bread prepared in advance.
- No assembled BeefLettuce, BeefBurgers, or BeefLettuceBurgers, but there is 1 LettuceBurger ready.
- 2 empty plates on counters.
- 1 fire extinguisher and no active fire.
- 18 empty counters.
- 2 orders pending: a BeefBurger with 30 seconds remaining and a LettuceBurger with 45 seconds remaining.
- Player 1 holding an empty plate.

Note that you will only receive the json below:

{GAME_STATE_EXAMPLE}

### Assigned Tasks

In this game, the `assigned tasks` are the actions and orders which you assign to the agent that need to **prioritize and complete urgently**.

Assigned tasks can be actions with pre-conditions, and order names.

#### Assigned Actions

Assigned tasks can contains pairs of preconditions and actions. Each pair specifies a condition that must be met and the corresponding action that should be taken when the condition is true. Here's a breakdown of what each element means:

1. **Precondition**:
    - A lambda function that takes `json_state` as an input and returns a boolean value.
    - It indicates whether a specific condition is met in the current game state.
    - For example: When you want to detect whether there are fewer than 3 well-cooked or in-process beefs, you can use `"lambda json_state: json_state['objects'][('Beef', 'Well-cooked')] + json_state['objects'][('Beef', 'In-progress')] < 3"`.

2. **Action**:
    - A tuple containing the action name and the action arguments.
    - The action name is a string, and the action arguments are provided as a dictionary.

#### Assigned Orders

The `assigned_tasks` can also contains the names of the orders that need to be completed in sequence.

- Each order (element) in this list is an order name in string.

#### Example Assigned Tasks

Now I will show you an example of assigned tasks below. In this example, the agent do the following tasks:
- prepare beef if the number of well-cooked or in-process beefs are fewer than the number of requirements.
- prepare a BeefBurger.
- prepare a LettuceBurger.

{ASSIGNED_TASKS_EXAMPLE}

Note that `assigned_tasks` will be executed in sequence **only once**, i.e., the actions will be executed if the preconditions are met and the orders will be prepared. If you want to prioritize some tasks, you can assign them in the head of `assigned_tasks`. Please pay attention to put the most urgent tasks in the head of the list.

**Urgent Needs**: `assigned_tasks` are mainly used for urgent needs you have found according to the latest (current) game state{LATEST_MESSAGE_PROMPT}.

# Examples

{FEW_SHOT_EXAMPLE}

# Input

{INPUT}


\end{lstlisting}

\subsection{Prompts of \framework}\label{app:dpt-prompt}
\subsubsection{Prompts of Code-as-Policy Generator}\label{app:cap-prompt}

\begin{lstlisting}
# OutputFormat

Please output in the following template:

You should return a text code block as your thought about how to prepare and serve burgers effectively.
```text
Be concise and clear, less than 50 words.
If no urgent responses are needed, return "Things are going well".
Do not directly copy the previous thoughts.
```

{MESSAGE_OUTPUT_FORMAT}

Return a **json** code block representation of the new assigned tasks that the agent will do urgently.
```json
**Pay attention that the agent will automatically prepare the burger order with the least remaining time and you should only assign tasks when changes are necessary.**
You can either keep some of the current assigned tasks if you find them still necessary, or substitute the current assigned tasks with the new ones, i.e., you don't need to include the current assigned tasks in the output.
You should make sure that the completed burgers are served to the customers in time, by letting the agent perform in default mode or adding serving actions. But do not serve the burgers that are not in the order list.
You should return an empty list (`[]`) here when the agent can automatically finish the orders itself and not urgent responses are needed.
Be careful to write correct lambda functions.
Do not directly copy the previous assigned tasks.
The JSON will be used in Python as `eval(json_string)`, so make sure it is in the correct format, e.g., use `True` and `False` instead of `true` and `false`.
```
\end{lstlisting}

\subsubsection{Prompts of Policy Reflection}\label{app:pr-promt}

\begin{lstlisting}
# OutputFormat

Please output in parts and in the following template:

You should return new **Behavior Guidelines** in the following code block.
```text
Analyze the past game history and identify areas for improvement or successful strategies. Then explain how the agent's policy will be adjusted based on the reflection.
Here are some suggestions for writing guidelines:
- What leads to the lost of scores, e.g., missed orders and served wrong food, in the past game?
- What leads to the waste of time in the past game?
- How to adjust the agent's policy to save time?
- What are the successful strategies in the past game?
- How to coordinate with the human player to achieve a higher score?
- How the agent's policy should be adjusted to improve performance?
- Why the beef is overcooked? How to avoid overcooking beef?
- Other suggestions for improving the performance of the team.
The guidelines should be given **based on the game history**.
You should return a text code block. Be concise and clear, less than 100 words.
```
\end{lstlisting}

\subsubsection{Prompts of Theory of Mind Module}\label{app:tom-prompt}

\begin{lstlisting}
You should return new **inference on the human player's behavior pattern** in the following code block.
```text
Analyze the past game history and identify patterns or tendencies in the human player's behaviors. Then explain how the agent's policy will be adjusted to coordinate better with the human player.
Here are some suggestions for writing inference:
- What are the human player's preferences in completing orders? For example, whether the human player prefers to complete orders with the least remaining time or orders with the most remaining time? This will help you determine which orders you should focus on to avoid missing any order and to prevent making extra food.
- How does the human player prioritize tasks when multiple orders are pending? For example, whether the human player tends to do order by order or tries to complete multiple orders simultaneously by preparing multiple ingredients in parallel?
- Which processes does the human player prefer to complete first? For example, whether the human player prefers to prepare which ingredients, assemble burgers or serve burgers? This will affect your choices regarding which tasks to prioritize. For example, when human player prefers to preparing ingredients, you choose to serve more dishes can effectively improve team efficiency. When human player prefers to assemble burgers, you can choose to prepare more ingredients and pass on to the counter to meet the requirements of the human player.
- Consider whether there are any patterns between human player's behavior, the current orders that need to be completed, and the ingredients available on the field. For example, humans tend to prepare a large amount of beef when multiple orders for beef burgers are needed. Such implicit patterns can help you adjust your own behavior.
- How the agent's policy should be adjusted to improve performance? For example, if you believe the ingredients you've prepared or the burgers you've made are meant for the human player to assemble or serve, you should pass them to the counter to facilitate efficient collaboration.
The inference should be given **based on the game history**.
You should return a text code block. Be concise and clear, less than 100 words.
```
\end{lstlisting}

\section{Implementation of Act, ReAct and Reflexion Frameworks}\label{app:baselines}
ReAct \cite{yao2022react} is a framework that integrates reasoning and acting by allowing agents to plan, interpret environments, and interact dynamically to improve decision-making.
Reflexion \cite{shinn2024reflexion} is a framework that enhances language model agents by enabling self-reflection, allowing them to learn from past mistakes, refine their reasoning, and iteratively improve decision-making in future tasks.
We use Act to name the LLM as Indenpendent System 1.

We implement Act, ReAct and Reflexion in Overcooked challenge. The three frameworks use the same prompt in the instruction part with \framework in \Cref{app:instruct}. We outline the specific differences in the output prompts for the three frameworks below.

\subsection{Output Prompts of Act}\label{app:act-prompt}
\begin{lstlisting}
# OutputFormat

Based on the current game state, considering the remaining time for the orders and the status of all ingredients on the kitchen, decide your next action.
Note that your actions should help advance the orders you're working on and the game process. Be sure to also consider your previous actions and their outcomes.
Please output a valid action in JSON format.
\end{lstlisting}

\subsection{Output Prompts of ReAct}\label{app:react-prompt}

\begin{lstlisting}
# OutputFormat

Please output in the following template:

You should return a text code block as your thought about how to prepare and serve burgers effectively.
```text
Be concise and clear, less than 50 words.
If no urgent responses are needed, return "Things are going well".
Do not directly copy the previous thoughts.
```

Your action should be a **json** code block representation of the new assigned tasks that the agent will do urgently.

```json
You can either keep some of the current assigned tasks if you find them still necessary, or substitute them with the new ones, i.e., you don't have to include the current assigned tasks in the output.
You should make sure that the completed burgers are served to the customers in time, by adding serving actions. But do not serve the burgers that are not in the order list.
You should return enough assigned tasks to keep the agent busy.
Be careful to write correct lambda functions.
Do not directly copy the previous assigned tasks.
The JSON will be used in Python as `eval(json_string)`, so make sure it is in the correct format, e.g., use `True` and `False` instead of `true` and `false`.
```
\end{lstlisting}

\subsection{Output Prompts of Reflexion}\label{app:refl-prompt}

\begin{lstlisting}
# OutputFormat

Based on a previous reasoning, you should improve based on self refection. Diagnose a possible reason for failure and devise a new, concise, high level plan that aims to mitigate the same failure. Use complete sentences.
You should return a text code block as your reflection when you meet the following failure situations: 1)Fire, 2)Missing Order, 3)Loss Score, 4)Other unexpected situations.
```text
Be concise and clear, less than 100 words.
If no reflection is needed, return "Things are going well".
Do not directly copy the previous reflection.
```
\end{lstlisting}

\twocolumn
\begin{table*}[t]
    \renewcommand{\arraystretch}{1.5} % 行高
    \centering
    \resizebox{\linewidth}{!}{
    \begin{tabular}{l|l}
    \toprule
    \textbf{Key Events}  & \textbf{Key Actions} \\
    \midrule
    Cook Beef & \textcircled{1} Get Beef from station Put onto Pan \\
    Use Beef & \textcircled{1} Plate well-done Beef from Pan \\
    Prepare Lettuce & \textcircled{1} Get lettuce from station \textcircled{2} Put onto Cutboard \textcircled{3} Chop Lettuce \\
    \multirow{2}{*}{Use Lettuce}      & 
                                        \textcircled{1} Plate Lettuce Done from Cutboard \textcircled{2} Plate Lettuce Done from Counter \textcircled{3} Put onto Plate with BeefBurger                     
                                        \\
                                        &
                                        \textcircled{4} Put onto Plate with Bread \textcircled{5} Put Lettuce onto Plate \textcircled{6} Put Lettuce onto Plate with Beef
                                        \\
    \multirow{2}{*}{Use Bread} &         
                                        \textcircled{1} Get Bread from Station \textcircled{2} Plate Bread from Counter \textcircled{3} Put onto Plate with BeefLettuce
                                        \\
                                        & 
                                        \textcircled{4} Put onto Plate with Lettuce \textcircled{5} Put Bread onto Plate \textcircled{6} Put Bread onto Plate with Beef
                                        \\

         Use Plate & \textcircled{1} Get Plate from Station \\
                  Serve & \textcircled{1} Deliver Burger \\
    \bottomrule
    \end{tabular}

    }
    \caption{\textbf{The mapping from key event to key actions from \citet{zhang2024mutual}.}}\label{tab:event_action}
    
\end{table*}

\section{Metrics}\label{app:metrics}

Metrics we used in experiments include Atom Action Occupy,  Failure Missed, Failure Wrong Serve, Score Efficiency, Agent Contribution Rate, the total game score, and latency in second.

\paragraph{Atom Action Occupy.} The percentage of total time spent by in-game agents performing actions. $t_{atomic}$ refers to the number of time steps that have atomic action. $t_{total}$ refers to the total number of time steps.
\begin{equation}
    \text{Atom Action Occupy} = \frac{t_{atomic}}{t_{total}} 
\end{equation}

\paragraph{Failure Missed.} The number of orders missed of each games.

\paragraph{Failure Wrong Serve} The number of incorrect orders made by agents.

\paragraph{Score Efficiency.} The average score gained per macro action being executed.
$S_{total\_gain}$ refers to score gained and is excluding penalty points.
$\text{MA}_e$ refers to the number of macro action (MA) being executed.
\begin{equation}
    \text{Score Efficiency} = \frac{S_{total\_gain}}{\text{MA}_e}
\end{equation}

\paragraph{Latency.} The time in second that from the request to the output of a macro action or a code-as-policy output.

\paragraph{Agent Contribution Rate.} A concept from \citet{zhang2024mutual} to demonstrate the agent's contribution in each order based on the overcook environment. 

Below are the definition from \citet{zhang2024mutual}:
Key task events $KE$ are defined to track which team member completes specific tasks in Overcooked. Based on the burger-making process, each of the three burger types involves a set of essential, non-repeatable events. For instance, preparing a BeefBurger requires completing five key events: Cooking Beef, Using Beef, Using Bread, Using a Plate, and Serving. Each of these key events occurs only once. The completion of these events is triggered by specific “interact” actions, which are referred to as Key Actions.
The key actions mapping with the key events are in \Cref{tab:event_action}.
Each key event completed by a player is counted once as their contribution to the overall performance. 
Since these key events are non-repeatable, we can determine each player's contribution by tracking the key events they complete while preparing each successfully served burger.
We define the agent's contribution ratio $CR^{i}$ as:
$CR^{i} = \frac{KE^{i}}{KE^{i}+KE^{h}} \times 100\%$, where $KE^{i}$ and $KE^{h}$ represent the key events completed by the agent and the human respectively.

\section{Details of the LLM as Independent \textit{System 1} and \textit{System 2} Experiments .}\label{app:exp-p}

\subsection{Models and Deployment}
In this series of experiments, we use 8 different model series including GPT \cite{openai2024gpt4omini}, Qwen \cite{qwen2.5}, Llama \cite{touvron2023llama}, Phi \cite{abdin2024phi4}, Gemma \cite{team2024gemma}, Mistral \cite{mistralAI}, DeepSeek \cite{liu2024deepseek} and DeepSeek-R1 \cite{guo2025deepseek}:

\textbf{GPT Series}: GPT-4o, GPT-4o-mini and o3-mini

 \textbf{Qwen Series}: Qwen2.5 with 5 different sizes including 3b, 7b, 14b, 32b and 72b (Lisence: Apache license 2.0)
 
\textbf{Llama Series}: Llama3.1-8b, Llama3.2-3b and Llama3.3-70b (Lisence: llama)

\textbf{Phi Series}: Phi-3.5-3.8b and Phi-4-14b (Lisence: MIT)

\textbf{Gemma Series}: Gemma2 with 3 different sizes including 2b, 9b and 27b (Lisence: gemma)

\textbf{Mistral Series}: Ministral with 2 different sizes including 3b and 8b, Mistral-nemo-12b, Mistral-small-24b and Mixtral-8x22b (Lisence: mistral)

\textbf{DeepSeek Series}: DeepSeek-V2-16b and DeepSeek-V2.5 (Lisence: MIT)

\textbf{DeepSeek-R1 Series}: DeepSeek-R1 with 5 different sizes including 7b, 8b, 14b, 32b and 70b (Lisence: MIT)

All the open-source models are locally deployed with NVIDIA A800-SXM4-80GB through ollama \cite{ollama}, with the number of cards used determined by the model size.
For DeepSeek-R1 series in Long CoT, we deploy via llama.cpp \cite{llama.cpp} for customizing structured output.
The GPT series models use native API calls to conduct experiments.
The experiments use 26.3 A800-SXM4-80GB GPU hours for open-source models and \$ 35 in OpenAI API cost.
All models had their temperature parameter set to 0, while the remaining parameters were kept at their default values.

\subsection{Detailed Results}

We list the data from \Cref{fig:exp1} in \Cref{tab:exp1-1} and provided more detailed metrics. 
Metrics include Atom Action Occupy, Failure Missed, Failure Wrong Serve), Score Efficiency, the total game score, and latency in second.

\section{Details of Capability in Real-time Task Experiments .} \label{app:exp1}

\subsection{Models and Deployment}

In this series of experiments, we used 5 different model series including GPT \cite{openai2024gpt4omini}, Qwen \cite{qwen2.5}, Llama \cite{touvron2023llama}, Mistral \cite{mistralAI} and DeepSeek \cite{guo2025deepseek}.

\textbf{GPT Series}: GPT-4o, GPT-4o-mini and o3-mini

\textbf{Qwen Series}: Qwen2.5 with 3 different sizes including 14b, 32b and 72b (Lisence: Apache license 2.0)

\textbf{Llama Series}: Llama3.3-70b  (Lisence: llama)

\textbf{Mistral Series}: Mistral-nemo-12b, Mistral-small-24b and Mixtral-8x22b (Lisence: mistral)

\textbf{DeepSeek Series}: DeepSeek-R1-Distill-Llama-70B and DeepSeek-V2.5 (Lisence: MIT)

All the open-source models are locally deployed with NVIDIA A800-SXM4-80GB through vLLM \cite{kwon2023efficient}, with the number of cards used determined by the model size.
For DeepSeek-R1-70b, we use 8 NVIDIA H100-80GB-HBM3 for deployment through vLLM \cite{kwon2023efficient}.
The GPT series models use native API calls to conduct experiments.
The experiments use 140.3 A800-SXM4-80GB GPU hours and 17.5 H100-80GB-HBM3 GPU hours for open-source models and \$ 100 in OpenAI API cost.
All models had their temperature parameter set to 0, while the remaining parameters were kept at their default values.

\subsection{Detailed Results}

We list the data from \Cref{fig:exp1-2} in \Cref{tab:exp1-2-react,tab:exp1-2-reflexion,tab:exp1-2-dpt} and provided more detailed metrics. 
Metrics include Atom Action Occupy, Failure Missed, Failure Wrong Serve), Score Efficiency, the total game score, and latency in second.

\onecolumn
\begin{sidewaystable}
    \centering
    \resizebox{0.6\paperheight}{!}{
    \begin{tabular}{lcccccc}
    \toprule
\multicolumn{1}{c}{\multirow{2}{*}{\diagbox[width=3.5cm]{\textbf{Model}}{\textbf{Metrics}}}} & \textbf{Atom Action Occupy } & \textbf{Failure Missed} & \textbf{Failure Wrong Serve}& \textbf{Score Efficiency} & \textbf{Score} & \textbf{Latency}\\
& - & Times & Times&  Score/Marco Action&  &  Second \\
\midrule
\textbf{FSM}  & \textbf{0.90}  &3.00 & \textbf{0.00}  & \textbf{5.59}  & \textbf{65.00}  & \textbf{0.00(0.00)} \\\midrule
\midrule
\multicolumn{7}{c}{LLM as Independent \textit{System 1}}\\
\midrule
\textbf{GPT-4o}  & 0.83(0.00)  & \textbf{2.70(0.18)}  & 0.00(0.05)  & \uline{2.00(0.10)}  & 46.00(4.00)  & 0.91(0.03) \\
\textbf{GPT-4o-mini}  & 0.85(0.01)  & 3.40(0.22)  & 0.00(0.10)  & 1.43(0.10)  & 24.50(5.97)  & 0.78(0.03) \\
\midrule
\textbf{DeepSeek-V2-16b}  & 0.00(0.00)  & 4.00(0.00)  & \textbf{0.00(0.00)}  & 0.00(0.00)  & -40.00(0.00)  & 1.62(0.00) \\
\textbf{DeepSeek-V2.5-236b}  & 0.52(0.00)  & 3.00(0.00)  & \textbf{0.00(0.00)}  & 1.39(0.01)  & -5.00(0.00)  & 6.51(0.03) \\
\midrule
\textbf{Gemma2-2b}  & \uline{0.00(0.00)}  & 4.00(0.00)  & \textbf{0.00(0.00)}  & 0.00(0.00)  & -40.00(0.00)  & 0.92(0.01) \\
\textbf{Gemma2-9b}  & 0.77(0.00)  & 4.10(0.11)  & 0.10(0.11)  & 0.49(0.06)  & -24.00(4.25)  &1.44(0.01) \\
\textbf{Gemma2-27b}  & 0.74(0.00)  & 4.00(0.09)  & {1.00(0.07)}  & 0.74(0.06)  & -30.00(2.56)  & 2.26(0.02) \\
\midrule
\textbf{Llama3.1-8b } & 0.82(0.00)  & 4.30(0.11)  & \textbf{0.00(0.00)}  & 0.00(0.00)  & -43.00(1.12)  & 1.07(0.02) \\
\textbf{Llama3.2-3b}  & 0.80(0.00)  & 4.00(0.00)  & \textbf{0.00(0.00)}  & 0.00(0.00)  & -40.00(0.00)  & 0.64(0.01) \\
\textbf{Llama3.3-70b} & 0.54(0.00)  & 4.00(0.00)  & 4.00(0.00)  & 1.40(0.3)  & -5.00(0.75)  & 4.38(0.02) \\
\midrule
\textbf{Ministral-3b}  & 0.00(0.00)  & 4.00(0.00)  & \textbf{0.00(0.00)}  & 0.00(0.00)  & -40.00(0.00)  & \uline{0.62(0.00)} \\
\textbf{Ministral-8b}  & 0.75(0.03)  & 5.00(0.08)  & \textbf{0.00(0.00)}  & 0.64(0.05)  & -25.00(1.06)  & 1.03(3.15) \\
\textbf{Mistral-nemo-12b}  & 0.72(0.02)  & 4.00(0.05)  & {1.00(0.08)}  & 1.27(0.11)  & 2.50(2.55)  & 1.29(0.33) \\
\textbf{Mistral-small-24b} & 0.70(0.01)  & 5.30(0.21)  & {0.50(0.11)}  & 0.31(0.10)  & -47.50(5.57)  & 1.86(0.04) \\
\textbf{Mixtral-8x22b}& 0.72(0.00)  & 4.00(0.11)  & {0.40(0.11)}  & 1.27(0.11)  & -12.50(4.20)  & 2.38(0.02) \\
\midrule
\textbf{Phi-3.5-3.8b}  & 0.75(0.00)  & 4.00(0.09)  & \textbf{0.00(0.00)}  & 0.00(0.00)  & -40.00(0.92)  & 1.00(0.03) \\
\textbf{Phi-4-14b}  & 0.81(0.00)  & 5.00(0.55)  & \textbf{0.00(0.00)}  & 0.00(0.03)  & -500.00(1.00)  & 1.51(0.01) \\
\midrule
\textbf{Qwen2.5-3b}  & 0.93(0.03)  & 5.00(0.00)  & \textbf{0.00(0.00)}  & 0.00(0.00)  & -50.00(0.00)  & 0.90(0.29) \\
\textbf{Qwen2.5-7b}  & 0.78(0.03)  & 5.00(0.45)  & \textbf{0.00(0.00)}  & 0.00(0.00)  & -50.00(4.47)  & 2.10(0.48) \\
\textbf{Qwen2.5-14b}  & 0.76(0.00)  & 3.50(0.11)  & {0.00(0.10)}  & \uline{2.29(0.12)}  & 31.50(4.21)  & 1.59(0.03) \\
\textbf{Qwen2.5-32b}  & 0.63(0.02)  & 3.00(0.12)  &\textbf{0.00(0.00)}  & 1.59(0.07)  & 18.00(3.62)  & 2.67(0.52) \\
\textbf{Qwen2.5-72b}  & 0.56(0.00)  & 4.00(0.24)  & 0.90(0.11)  &0.16(0.18)  & -45.00(5.26)  & 4.56(0.04) \\
\midrule
\multicolumn{7}{c}{LLM as Independent \textit{System 2}}\\
\midrule
\textbf{o3-mini}  & 0.51(0.01)  & 4.00(0.10)  & {0.00(0.09)}  & 0.94(0.13)  & -20.50(3.82)  & 5.10(0.30) \\
\midrule
\textbf{DeepSeek-R1-7b} & 0.00(0.00)  & 4.00(0.00)  & \textbf{0.00(0.00)}  & 0.00(0.00)  & -40.00(0.00)  & 1.38(0.01) \\
\textbf{DeepSeek-R1-8b} & 0.00(0.00)  & 4.00(0.05)  & \textbf{0.00(0.00)}  & 0.00(0.03)  & -40.00(2.23)  & 1.36(0.01) \\
\textbf{DeepSeek-R1-14b} & 0.67(0.02)  & 4.00(0.11)  & {1.80(0.29)}  & 0.47(0.10)  & -45.00(6.68)  & 2.20(0.01) \\
\textbf{DeepSeek-R1-32b}& 0.54(0.01)  & 3.10(0.11)  & {0.10(0.11)}  & 0.80(0.12)  & -13.00(4.19)  & 3.86(0.02) \\
\textbf{DeepSeek-R1-70b} & 0.45(0.00)  & 5.00(0.07)  & \textbf{0.00(0.00)}  & 0.38(0.10)  & -42.50(2.31)  & 6.69(0.03) \\
\midrule
\textbf{QwQ-32b} & 0.55(0.00)  & 4.00(0.00)  & \textbf{0.00(0.00)}  & 1.46(0.05)  & -5.00(1.03)  & 4.54(0.01) \\
        \bottomrule
    \end{tabular}
}
    \caption{Performance of Different Models in LLM as Independent \textit{System 1} and \textit{System 2} Experiments.}
    \label{tab:exp1-1}
\end{sidewaystable}

\clearpage
\twocolumn
\begin{sidewaystable}
    \centering
    \resizebox{0.55\paperheight}{!}{
    \begin{tabular}{lcccccc}
        \toprule
\multicolumn{1}{c}{\multirow{2}{*}{\diagbox[width=3.5cm]{\textbf{Model}}{\textbf{Metrics}}}} & \textbf{Atom Action Occupy } & \textbf{Failure Missed} & \textbf{Failure Wrong Serve}& \textbf{Score Efficiency} & \textbf{Score} & \textbf{Latency}\\
        & - & Time & Time&  Score/Marco Action&  &  Second \\
        \midrule
 \textbf{GPT-4o}  & 0.87(0.02)  & 3.70(0.15)  & 0.00(0.10)  & 3.08(0.30)  & 21.00(7.01)  & 7.10(0.29) \\
\textbf{GPT-4o-mini}  & 0.90(0.03)  & 4.00(0.12)  & 0.00(0.16) & 0.60(0.28)  & -28.50(6.23)  & 3.06(0.07) \\
 \textbf{o3-mini-low}  &    0.70(0.02) &  3.60(0.18) &0.00(0.08)  & 2.51(0.25)  &5.50(5.86)  & 8.64(0.27) \\
\midrule
\textbf{DeepSeek-V2.5-236b}  &  0.63(0.02) &  4.30(0.15) & 0.00(0.05)  &1.72(0.24)  &-21.50(3.56)  & 6.45(0.18)\\
\textbf{DeepSeek-R1-70b} &  0.67(0.02) & 4.00(0.10) & 0.70(0.11)  & 1.48(0.17)  &-17.00(4.32)  & 7.79(0.20)\\
\textbf{DeepSeek-R1-32b}   & 0.70(0.03)           & 4.00(0.13)  & 0.00(0.13) & 1.49(0.18) & -15.50(4.51)  & 5.77(0.18)  \\
\textbf{DeepSeek-R1-14b}   & 0.87(0.02)  & 4.00(0.12)  &0.00(0.12) & 2.67(0.19)  & -7.00(4.94)  & 2.91(0.03)  \\
\midrule
\textbf{Llama3.3-70b} &    0.82(0.01) &  4.00(0.12) & \textbf{0.00(0.00)}  & 2.86(0.16)  &20.00(4.21)  & 5.44(0.05)\\
\midrule
\textbf{Mistral-nemo-12b} &    0.64(0.01) &  4.50(0.17) & \textbf{0.00(0.00)}  & 2.40(0.13)  &-10.00(3.31)  & \textbf{1.10(0.03)}\\
\textbf{Mistral-small-24b}   &  \textbf{0.94(0.01)} &  3.00(0.11) & \textbf{0.00(0.00)}  & \textbf{4.63(0.20)}  &\textbf{59.50(5.04)}  & 2.69(0.02) \\
\textbf{Mixtral-8x22b}  &  0.88(0.02) & 3.40(0.15) & \textbf{0.00(0.00)}  & 1.73(0.22)  &-5.00(5.23)  & 5.56(0.10) \\
\midrule
\textbf{Qwen2.5-14b}   & 0.90(0.02)&  3.50(0.18) & 0.20(0.18)  & 1.98(0.21)  &-5.00(5.31)  & 1.55(0.03)\\
\textbf{Qwen2.5-32b}  & 0.87(0.00)  & 4.00(0.05)  & \textbf{0.00(0.00)}  & 2.94(0.02)  & 10.00(0.50)  & 1.93(0.04) \\
\textbf{Qwen2.5-72b}  &  0.83(0.03) &  \textbf{2.90(0.12)} & 0.00(0.05)  & 2.71(0.09)  &16.50(3.22)  & 4.60(0.09) \\
\textbf{QwQ-32b}  & 0.78(0.01)  & 3.20(0.11)  & \textbf{0.00(0.00)}  & 2.46(0.12)  & 8.00(2.77)  & 10.75(0.24) \\

        \bottomrule
    \end{tabular}}
    \caption{Performance of Different Models - ReAct }
    \label{tab:exp1-2-react}
\end{sidewaystable}

\begin{sidewaystable}
    \centering
    \resizebox{0.55\paperheight}{!}{
    \begin{tabular}{lcccccc}
\toprule
\multicolumn{1}{c}{\multirow{2}{*}{\diagbox[width=3.5cm]{\textbf{Model}}{\textbf{Metrics}}}} & \textbf{Atom Action Occupy } & \textbf{Failure Missed} & \textbf{Failure Wrong Serve}& \textbf{Score Efficiency} & \textbf{Score} & \textbf{Latency}\\
& - & Times & Times&  Score/Marco Action&  &  Second \\
\midrule
\textbf{GPT-4o}  & 0.80(0.02)  & 4.00(0.16)  & 0.00(0.09)  & 2.14(0.17)  & -1.50(3.78)  & 7.49(0.27) \\
\textbf{GPT-4o-mini}  & \textbf{0.91(0.01)}  & 4.00(0.07)  & \textbf{0.00(0.00)}  & 0.00(0.14)  & -40.00(2.17)  & 3.11(0.08) \\
\textbf{o3-mini}  &  0.78(0.02) &  4.00(0.18) & 0.30(0.11)  & 1.78(0.26)  &-16.50(7.12)  & 8.86(0.23) \\
\midrule
\textbf{DeepSeek-V2.5}   &  0.52(0.01) &  4.22(0.13) & \textbf{0.00(0.00)}  & 1.24(0.18)  &-25.56(2.91)  & 7.64(0.16)\\
\textbf{DeepSeek-R1-70b}  &  0.66(0.01) &  4.00(0.14) & 0.00(0.08)  & 1.44(0.19)  &-20.00(4.79)  & 7.78(0.17) \\
\textbf{DeepSeek-R1-32b}   & 0.79(0.02)           & 4.60(0.11)  & 0.30(0.17) & 0.90(0.21) & -37.50(4.77)  & 7.39(0.11)  \\
\textbf{DeepSeek-R1-14b}   & 0.80(0.03)           & 3.7(0.15)  &0.00(0.07) & 1.93(0.22)  & -10.50(4.12)  & 4.01(0.11)  \\

\midrule
\textbf{Llama3.3-70b}   &  0.65(0.02) &  \textbf{3.00(0.09)} & \textbf{0.00(0.00)}  & \textbf{3.25(0.19)}  &\textbf{20.00(4.47)}  & 5.20(0.06) \\
\midrule
\textbf{Mistral-nemo-12b}   &  0.00(0.00) &  4.00(0.00) & \textbf{0.00(0.00)}  & 0.00(0.00) & -40.00(0.00)  & \textbf{1.60(0.02)} \\
\textbf{Mistral-small-24b}   &  0.90(0.01) &  \textbf{3.00(0.09)} & 0.00(0.10)  & 1.43(0.03)  &-5.00(3.63)  & 3.11(0.05) \\
\textbf{Mixtral-8x22b}   &  0.84(0.02) &  3.80(0.18) & \textbf{0.00(0.00)}  & 2.44(0.20)  & 0.50(4.33)  & 5.58(0.23) \\
\midrule
\textbf{Qwen2.5-14b}   &  \textbf{0.91(0.02)} &  4.00(0.10) & 0.00(0.10)  & 2.44(0.24)  &-4.00(4.45)  & 1.87(0.05) \\
\textbf{Qwen2.5-32b}  & 0.00(0.00)  & 4.00(0.00)  & \textbf{0.00(0.00)}  & 0.00(0.00)  & -40.00(0.00)  & 2.93(0.05) \\
\textbf{Qwen2.5-72b}  & 0.69(0.05)  & \uline{5.00(0.09)}  & \textbf{0.00(0.00)}  & 1.47(0.09)  & -25.00(2.76)  & 4.66(0.05) \\
\textbf{QwQ-32b}  & 0.00(0.01)  & 5.00(0.00)  & \textbf{0.00(0.00)}  & 0.00(0.11)  & -50.00(0.75)  & 7.75(0.11) \\
\bottomrule
    \end{tabular}}
        \caption{Performance of Different Models - Reflexion}
    \label{tab:exp1-2-reflexion}
\end{sidewaystable}

\clearpage
  \onecolumn

\begin{sidewaystable}
    \centering
    \resizebox{0.7\paperheight}{!}{
    \begin{tabular}{lcccccc}
        \toprule
       \multicolumn{1}{c}{\multirow{2}{*}{\diagbox[width=3.5cm]{\textbf{Model}}{\textbf{Metrics}}}} & \textbf{Atom Action Occupy } & \textbf{Failure Missed} & \textbf{Failure Wrong Serve}& \textbf{Score Efficiency} & \textbf{Score} & \textbf{Latency}\\
        & - & Times & Times&  Score/Marco Action&  &  Second \\
        \midrule
\textbf{GPT-4o}            & 0.91(0.00)           & 3.60(0.14)  & 0.00(0.05) & 3.05(0.24)  & 20.50(5.41)  & 5.08(0.15) \\
\textbf{GPT-4o-mini}       & 0.93(0.00)           & 3.60(0.11)  & \textbf{0.00(0.00)} & 3.50(0.23)  & 21.00(4.47)  & 2.13(0.01) \\
\textbf{o3-mini}           & 0.90(0.00)           & 3.00(0.18)  & \textbf{0.00(0.00)} & 3.68(0.19)  & 37.50(4.81)  & 7.03(0.28) \\
\midrule
\textbf{DeepSeek-V2.5}     & 0.93(0.00)           & 3.00(0.11)  &\textbf{ 0.00(0.00)} & 3.40(0.14)  &31.50(3.40)  & 4.73(0.11) \\
\textbf{DeepSeek-R1-70b}   & 0.90(0.01)           & \textbf{2.30(0.17)}  & \textbf{0.00(0.00)} & \textbf{4.19(0.15)}  & \textbf{60.00(4.35)}  & 9.09(0.26)  \\
\textbf{DeepSeek-R1-32b}   & 0.93(0.00)           & 2.80(0.21)  & \textbf{0.00(0.00)} & 3.35(0.27) & 39.50(7.68)  & 6.58(0.25)  \\
\textbf{DeepSeek-R1-14b}   & 0.91(0.01)           & 3.40(0.15)  &0.00(0.50) & 3.04(0.21)  & 23.00(5.42)  & 3.87(0.07)  \\
\midrule
\textbf{Llama3.3-70b}      & 0.94(0.00)           & 4.00(0.13)  & 0.00(0.05) & 1.82(0.34)  & -10.00(6.46) & 2.28(0.10)  \\
\midrule
\textbf{Mistral-nemo-12b}  & 0.91(0.01)           & 3.30(0.13)  & \textbf{0.00(0.00)} & 3.49(0.21)  & 30.00(5.20)  & 1.31(0.03)  \\
\textbf{Mistral-small-24b} & 0.91(0.00)           & 4.00(0.09)  & \textbf{0.00(0.00)} & 2.05(0.17)  & -1.50(3.63)  & 3.61(0.31)  \\
\textbf{Mixtral-8x22b}     & \textbf{0.95(0.01)}  & 4.00(0.00)  & \textbf{0.00(0.00)} & 2.70(0.20)  & 0.00(15.00)  & 4.21(0.17) \\
\midrule
\textbf{Qwen2.5-14b}       & 0.94(0.00)  & 4.00(0.05)  & \textbf{0.00(0.00)} & 2.68(0.22)  & 1.50(4.11)  & \textbf{1.18(0.02)}  \\
\textbf{Qwen2.5-32b}       & 0.93(0.00)           & 4.00(0.13)  & \textbf{0.00(0.00)} & 2.26(0.13)  & 1.00(3.83)  & 1.65(0.03)  \\
\textbf{Qwen2.5-72b}       & 0.94(0.01)           & 3.70(0.18)  &\textbf{ 0.00(0.00)} & 2.66(0.21) & 11.00(4.88)  & 3.01(0.12)   \\
\textbf{QwQ-32b}       & 0.90(0.00)           & 2.70(0.17)  &\textbf{ 0.00(0.00)} & 3.90(0.15) & 51.00(4.74)  & 14.96(0.78)   \\
        \bottomrule
    \end{tabular}
    }
        \caption{Performance of Different Models - \framework.}
    \label{tab:exp1-2-dpt}
\end{sidewaystable}

\clearpage
\twocolumn

\begin{table*}[htp]
\resizebox{\linewidth}{!}{
\begin{tabular}{lcccccccc}
\toprule
\multicolumn{1}{c}{\multirow{3}{*}{\diagbox[width=3.5cm]{\textbf{Model}}{\textbf{Framework}}}} &
  \multicolumn{4}{c}{\textbf{Score Efficiency}} &
  \multicolumn{4}{c}{\textbf{Latency}} \\
  \cmidrule(lr){2-5}
  \cmidrule(lr){6-9}
  
\multicolumn{1}{c}{}                               & \multirow{2}{*}{ReAct}    & \multirow{2}{*}{Reflexion}       & \framework    & \multirow{2}{*}{\framework}     & \multirow{2}{*}{ReAct}   & \multirow{2}{*}{Reflexion}     & \framework    & \multirow{2}{*}{\framework}   \\
\multicolumn{1}{c}{}           &     &        & w/o ToM   &   &    &      &  w/o ToM   &    \\
\midrule
\textbf{o3-mini-high} & 0.00(0.00)	&0.00(0.00)&	\textbf{5.66(0.21)}	&5.33(0.18) &39.01(2.82)	&39.47(2.43)	&\textbf{34.77(4.37)}&35.96(4.91)\\
\textbf{o3-mini-medium} & 2.67(0.38)	&3.59(0.39)	&5.16(0.28)&\textbf{5.23(0.24)} &28.07(2.42)	&26.73(3.93)	&\textbf{22.24(1.39)}&24.05(2.81)\\
\textbf{o3-mini-low} & 3.20(0.34)	&4.18(0.34)&	\textbf{4.28(0.43)}	&4.60(0.35) &10.78(1.40)	&10.58(0.80)	&\textbf{7.34(0.37)}&7.68(0.38)\\
\textbf{GPT-4o }  & 4.26(0.42)	&3.86(0.34)	&3.43(0.42)&	\textbf{4.46(0.39)}&	6.63(7.53)	&6.81(0.24)&	4.92(1.32)	&\textbf{4.91(1.41)}\\
\textbf{GPT-4o-mini}   & 3.95(0.52)	&4.64(0.66)	&5.03(0.28)&	\textbf{5.33(0.33)} &  2.93(0.77)	&3.15(1.27)	&2.09(1.09)	
&\textbf{2.08(0.58)} \\
\textbf{Qwen-Max}   &4.56(0.39)&	4.03(0.28)	&4.83(0.45)	&\textbf{5.09(0.31)} &8.29(0.14)&	10.30(0.21)&	5.90(0.11)&	\textbf{5.89(0.10)} \\
\textbf{Claude 3.5 Haiku}    & 4.04(0.30)	&3.65(0.31)&	\textbf{4.67(0.39)}&	4.47(0.34)  &5.74(0.06)	&7.47(0.11)&	\textbf{5.21(0.05)}&	5.25(0.06)\\
% \textbf{Claude 3.5 Sonnet}  &  & -& - & - & - &  &- & - & - & -  \\
% \textbf{Gemini 2.0 Flash}  &  \textbf{58.33}   & -& - & - & - & \textbf{0.67}  &- & - & - & -   \\
\midrule
\textbf{DeepSeek-R1-671b}  &4.52(0.25)	&4.47(0.37)
&	4.90(0.21)	&\textbf{5.27(0.19)}	 &\textbf{ 31.31(2.17)}	&41.66(2.45)&	38.89(3.70)	&34.63(2.30)\\
\textbf{DeepSeek-R1-70b}  &3.66(0.25)&	2.25(0.27)	&4.64(0.25)&	\textbf{4.92(0.24)} &7.82(0.17)&	\textbf{7.39(0.14)}	&10.30(0.36)&	10.13(0.34)\\
\textbf{DeepSeek-R1-32b}  &3.64(0.29)&	3.27(0.29)	&\textbf{4.31(0.26)}&	4.04(0.38) &5.75(0.09)	&6.77(0.13)&	5.24(0.08)&	\textbf{5.11(0.13)} \\
\textbf{DeepSeek-R1-14b}  &3.16(0.22)&	3.16(0.43)	&\textbf{4.33(0.36)}	&4.29(0.38) & \textbf{3.06(0.06)}	&3.44(0.09)&	3.88(0.088)	&3.57(0.06)\\
\textbf{DeepSeek-V3 }   &4.78(0.39)&	5.03(0.38)	&\textbf{6.00(0.18)}	&5.66(0.25)&   7.54(0.15)&8.86(0.15)	&\textbf{1.92(0.04)}	&2.41(0.10)  \\
\textbf{DeepSeek-V2.5}  &	2.29(0.26)	&3.43(0.29)	&\textbf{4.24(0.40)	}&3.61(0.42)&  4.88(0.07)	&5.35(0.08)	&\textbf{4.06(0.10)	}&4.49(0.07)	\\
\textbf{QwQ-32b} & 3.92(0.24)	&0.00(0.21)&\textbf{4.64(0.25)}	&4.29(0.25)	&	8.80(1.04)	&\textbf{7.28(2.47)}&14.50(3.04)	&12.67(1.21)	 \\
\textbf{Qwen2.5-72b} & 4.44(0.16)&	\textbf{5.11(0.29)}	&3.25(0.27)	&4.51(0.29)&	4.34(0.06)	&4.83(0.11)	&\textbf{3.81(0.10)}	&4.62(0.11) \\
\textbf{Llama3.3-70b}  & 4.44(0.37)	&2.01(0.28)	&\textbf{4.08(0.19)}&	3.89(0.32)&    4.53(0.08)&	5.34(0.11)	&\textbf{2.30(0.08)}	&2.90(0.09) \\
\textbf{Mixtral-8x22b}   &  3.58(0.30)&	4.01(0.32)	&\textbf{4.63(0.41)}&	4.38(0.43)&  5.20(0.18)	&5.19(0.22)	&\textbf{4.53(0.14)}&	5.31(0.19)  \\
\midrule
\midrule
\textbf{Overall}     &3.59(0.30)&	3.33(0.32)&	4.59(0.31)	&\textbf{4.67(0.31)	}  	&10.86(1.13)&	11.80(0.88)&10.11(0.97)	&\textbf{10.10(0.87)}\\
\bottomrule
\end{tabular}
}
\caption{\textbf{Results with Standard Errors of Experiments - Collaborating with Rule-based Agents.}}\label{tab:exp2-d}
\end{table*}

\section{Details of Capability in Simultaneous Collaboration Experiments} \label{app:exp2}

\subsection{Models and Deployment}

In this series of experiments, we used 5 different model series including GPT \cite{openai2024gpt4omini}, Claude \cite{anthropic_claude35}, Qwen \cite{qwen2.5}, Llama \cite{touvron2023llama}, Mistral \cite{mistralAI} and DeepSeek \cite{guo2025deepseek}.

\textbf{GPT Series}: GPT-4o, GPT-4o-mini and o3-mini

\textbf{Claude Series}: Claude-3.5-haiku

\textbf{Qwen Series}: Qwen2.5-72b  (Lisence: Apache license 2.0) and Qwen-Max

\textbf{Llama Series}: Llama3.3-70b  (Lisence: llama)

\textbf{Mistral Series}: Mixtral-8x22b  (Lisence: mistral)

\textbf{DeepSeek Series}: DeepSeek-R1-671b, DeepSeek-V2.5 and DeepSeek-V3 (Lisence: MIT)

All the open-source models are locally deployed with NVIDIA A800-SXM4-80GB through vLLM \cite{kwon2023efficient}, with the number of cards used determined by the model size.
For DeepSeek-R1-70b and DeepSeek-V3, we use 8 and 16 NVIDIA H100 80GB HBM3 for deployment through vLLM \cite{kwon2023efficient}.
The GPT series, Claude-3.5-haiku and Qwen-Max use native API calls to conduct experiments.
The experiments use 350.9 A800-SXM4-80GB GPU hours and 210.5 H100-80GB-HBM3 GPU hours for open-source models, \$ 400 in OpenAI API cost, \$ 80 in Claude API cost and \$ 120 in Qwen API cost.
All models had their temperature parameter set to 0, while the remaining parameters were kept at their default values.

% \subsection{Additional Results}

\subsection{Other Metrics}
We list the more metrics \Cref{tab:exp2-d} including Score Efficiency and latency in second.

\subsection{Case Study}\label{app:case-study}

\subsubsection{Example of \framework when playing with Beef Agent.}
Below is the ToM module output of \framework in the 78th timestep, driven by DeepSeek-R1-70b. It demonstrates that \framework's ToM mechanism effectively helps the agent infer that its partner specializes in preparing beef.
\begin{lstlisting}
Inferred Human Behavior:
- The human player focuses on beef preparation and distribution, frequently passing well-cooked beef to the counter.
- The agent should complement this by handling lettuce preparation, bread management, and burger assembly.
- The human's consistent passing of beef indicates a preference for handling meat, so the agent should focus on other tasks to optimize teamwork.
- The agent should pass prepared lettuce and bread to the counter to enable the human to assemble burgers efficiently.
- By dividing tasks based on the human's focus, the agent can improve overall efficiency and ensure orders are completed on time.
\end{lstlisting}

\subsubsection{Example of \framework when playing with Assemble \& Serve Agent.}
Below is the Reflection and ToM module output of \framework in the 266th timestep, driven by o3-mini-low. It demonstrates that \framework's ToM mechanism effectively helps the agent infer that its partner specializes in burger assembly.
\begin{lstlisting}
Behavior Guidelines:
Focus on avoiding overcooking by closely monitoring beef cooking times and switching to fresh ingredients if needed. Prioritize orders with minimal remaining time to prevent missed orders. Avoid repetitive actions like serving beefburgers without verifying order demand. Delegate tasks by quickly preparing ingredients and passing them to the human player, minimizing counter clutter and idle time.

Inference on Human Player Behavior Pattern:
The human player prioritizes urgent beef-based orders and rapid assembly, often focusing on assembling and serving ready-to-go items. Their approach suggests a preference for quick, order-focused actions, emphasizing the need for prepped ingredients. The agent should support this by preparing well-cooked beef and promptly passing completed ingredients to ensure smoother coordination.
\end{lstlisting}

 \section{Details of Human Experiments }\label{app:humanexp}

\subsection{Procedure}
We recruited 71 participants from the university through its internal social platform. Each participant received a compensation of 50 RMB for their participation. To enhance engagement and attentiveness, we provided performance-based bonuses. Participants within each group were ranked based on their self-play performance and their performance across four different agent games. The top 25\% in playing with each agent and self-play can receive an additional bonus of 3 RMB with a maximum possible bonus of 15 RMB.

The experiment was conducted online, where participants completed the tasks on a designated webpage using a computer with a keyboard. Each session lasted approximately 40 minutes. Participants controlled the chef using the arrow keys and interacted with objects by pressing the spacebar. The entire experimental process was recorded, with playback support available for data validation.

Participants were randomly assigned to one of two maps. Within a group, each participant interacted with four different agents, playing two games per agent, resulting in a total of eight games in random order. To examine whether participants could infer the agents' capabilities, they were not informed of the agent types but were only made aware that the experiment involved four types of agents, differentiated by color.

\begin{figure*}
    \centering
    \includegraphics[width=0.8\linewidth]{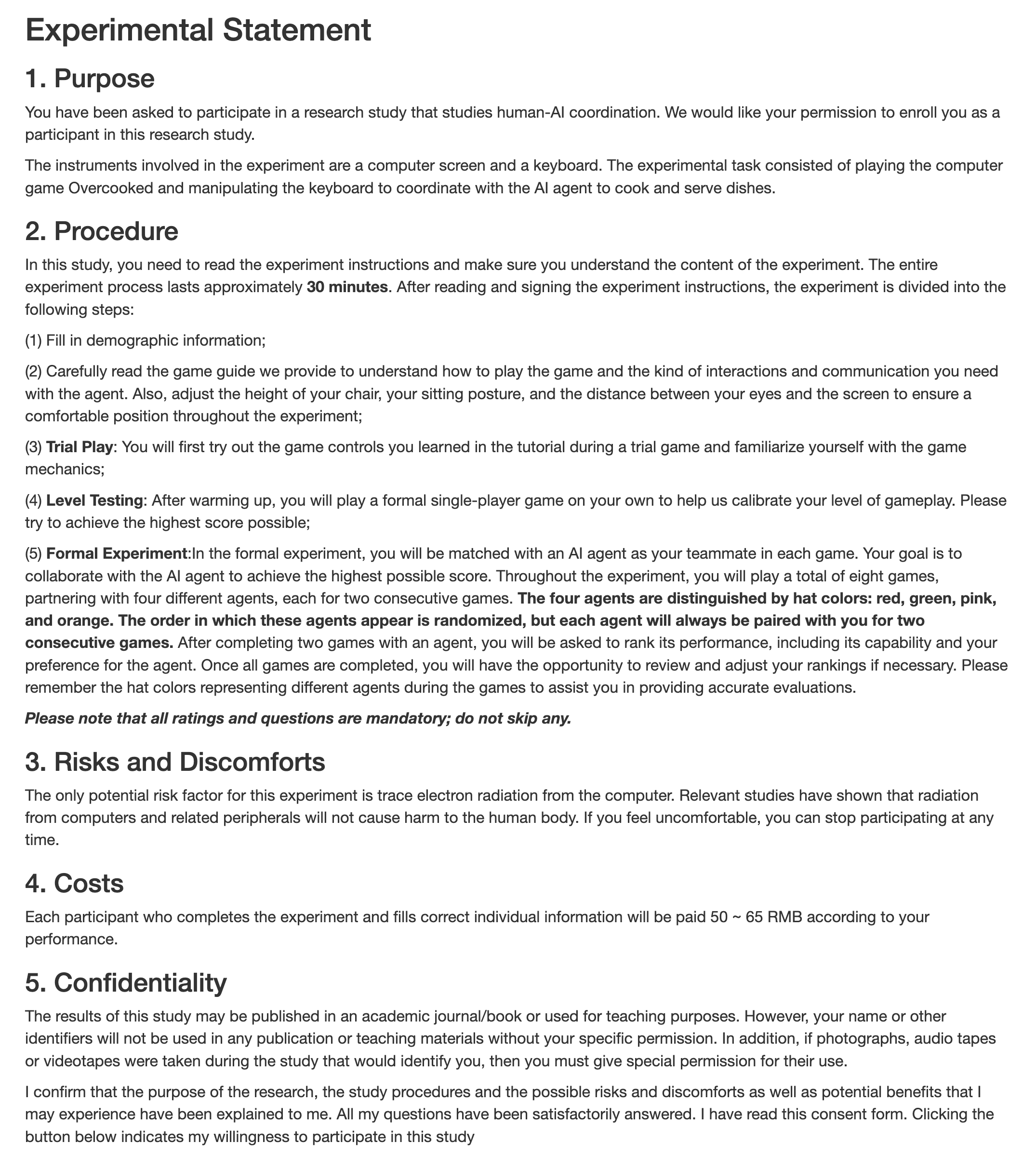}
    \caption{Experiment Statement.}
    \label{fig:statement}
\end{figure*}

Before beginning the experiment, all participants completed an informed consent form (\Cref{fig:statement}) and read instructions detailing the game rules and operations. Following the instructions, they first participated in a non-scored trial to familiarize themselves with the environment, rules, and controls. This was followed by a scored trial to assist with data validation.

\begin{figure*}
    \centering
    \subfigure[In Game]{\includegraphics[width=0.75\linewidth]{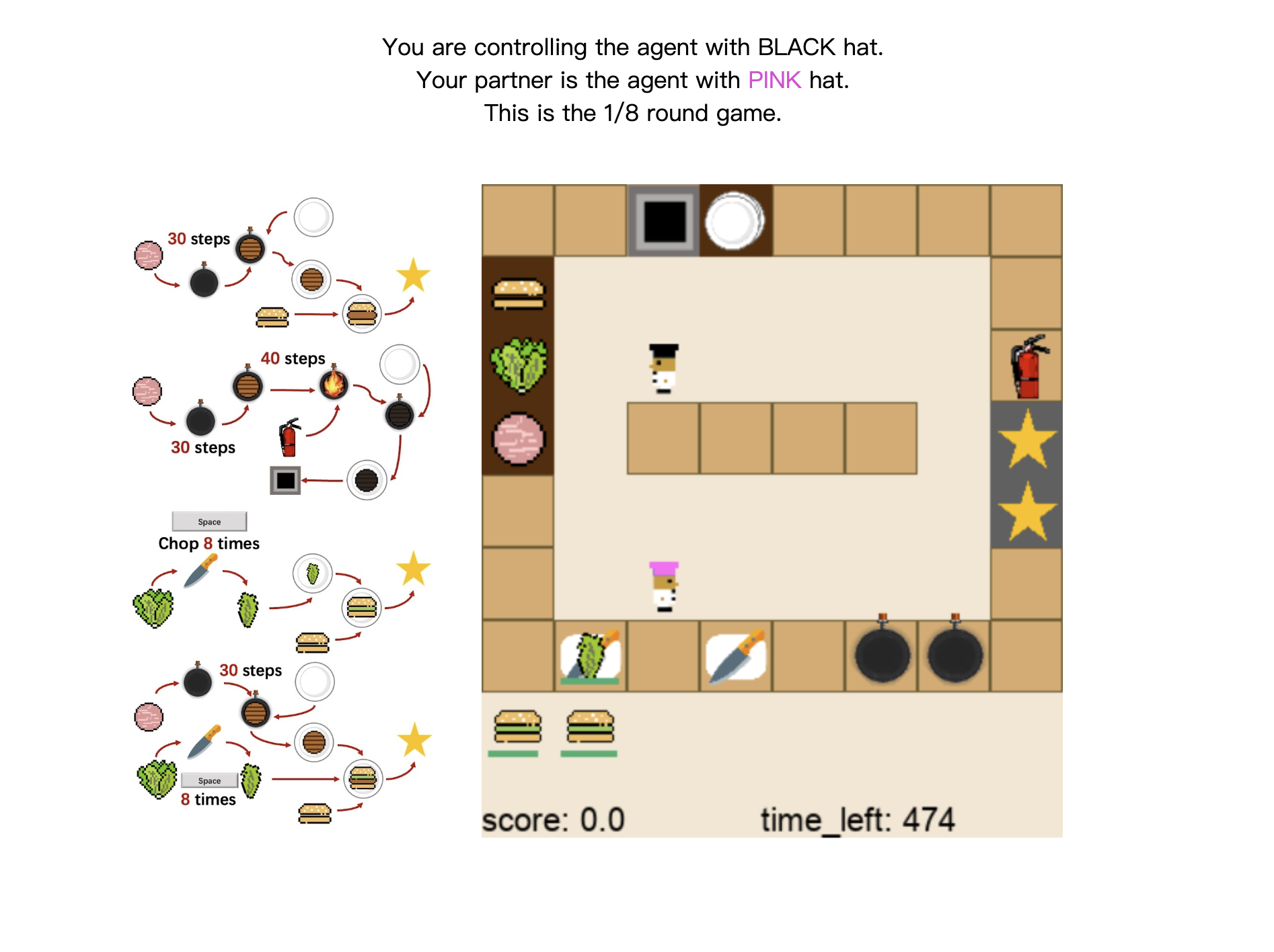}}
    \hfill
    \subfigure[Questionnaire]{\includegraphics[width=0.75\linewidth]{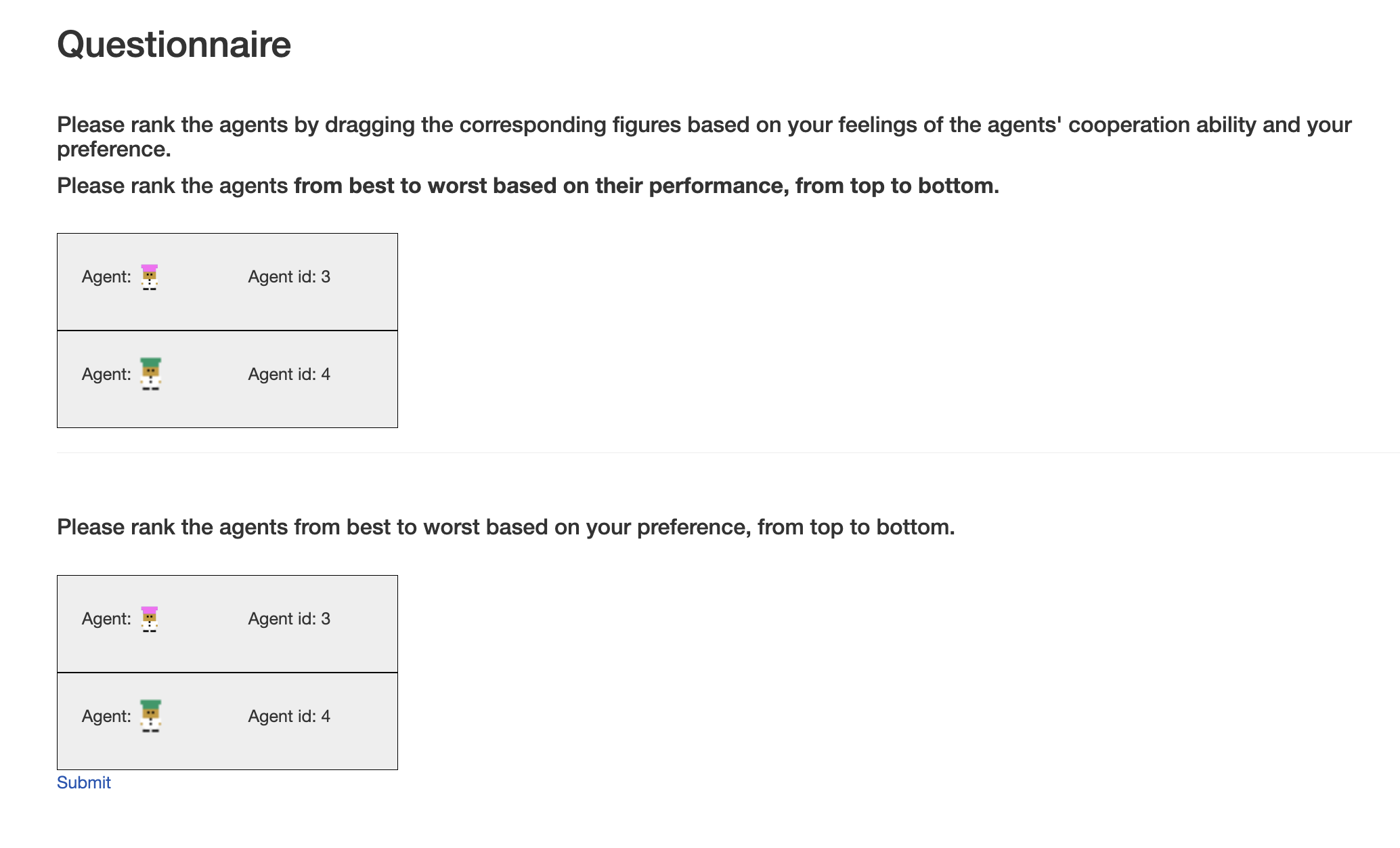}}
        \caption{\textbf{Experiment UI}}
    \label{fig:ui}
\end{figure*}

In the formal experiment (\Cref{fig:ui}), after each game, participants were asked to rank the agents based on their collaborative capabilities and personal preference. Upon completing all eight games, they filled out an additional questionnaire, where we collected their perceptions of task load and their intended level of task engagement.

The experiment use \$ 30 in OpenAI API cost.

\subsection{Participants}
A total of 71 participants participated in the study. 
The first and second authors of the article independently validated all collected data. 
This validation included checking data completeness (e.g., whether participants completed all the experiments) and reviewing the recorded playbacks to identify any abnormal actions (e.g., instances where participants did not engage in any cooperative behavior). 
After data validation, we excluded any data with anomalies, including passive participation and missing data, resulting in 68 valid participants (M = 36, F = 32, and Others = 0, ages between 18 and 31). Group 1 (Map 1) has 36 valid data points and Group 2 (Map 2) has 32 valid data points.

\end{document}